\documentclass[lettersize,journal]{IEEEtran}
\usepackage{amsmath,amsfonts}
\usepackage[algo2e]{algorithm2e}  
\usepackage{algorithm}             

\usepackage{array}
\usepackage{textcomp}
\usepackage{stfloats}
\usepackage{url}
\usepackage{verbatim}
\usepackage{graphicx}
\usepackage{cite}
\usepackage{subfigure}   
\usepackage{fancyhdr}
\usepackage{multirow}
\usepackage{graphicx}
\usepackage{amsfonts} 
\usepackage{float}
\usepackage{lipsum}
\usepackage{algpseudocode}
\usepackage{hyperref}
\usepackage{multirow}
\usepackage[table,xcdraw]{xcolor}

\begin{document}

\title{Image Super-Resolution Reconstruction Network\\ based on Enhanced Swin Transformer via Alternating Aggregation of Local-Global Features} 

\author{
	Yuming Huang$^{1,2}$, Yingpin Chen$^{2*}$, Changhui Wu$^2$, Binhui Song$^2$, Hui Wang$^2$
	
	{$^1$ School of Artificial Intelligence, Yulin Normal University}
	
	{$^2$School of Physics and Engineering, Minnan Normal University}
	\thanks{$^{*}$Corresponding author: Yingpin Chen (e-mail: 110500617@163.com).}
	\thanks{$^{*}$Co-first author: Yingpin Chen contribute equally to the first author.}
}

\markboth{Journal of \LaTeX\ Class Files,~Vol.~xx, No.~x, xxx~20xx}%
{xx \MakeLowercase{\textit{et al.}}: xx}


\maketitle

\thispagestyle{fancy}

\lfoot{\footnotesize © 2025 IEEE.  Personal use of this material is permitted.  Permission from IEEE must be obtained for all other uses, in any current or future media, including reprinting/republishing this material for advertising or promotional purposes, creating new collective works, for resale or redistribution to servers or lists, or reuse of any copyrighted component of this work in other works.}

\cfoot{}

\renewcommand{\headrulewidth}{0mm}

\begin{abstract}
The Swin Transformer image super-resolution (SR) reconstruction network primarily depends on the long-range relationship of the window and shifted window attention to explore features. However, this approach focuses only on global features, ignoring local ones, and considers only spatial interactions, disregarding channel and spatial-channel feature interactions, limiting its nonlinear mapping capability. Therefore, this study proposes an enhanced Swin Transformer network (ESTN) that alternately aggregates local and global features. During local feature aggregation, shift convolution facilitates the interaction between local spatial and channel information. During global feature aggregation, a block sparse global perception module is introduced, wherein spatial information is reorganized and the recombined features are then processed by a dense layer to achieve global perception. Additionally, multiscale self-attention and low-parameter residual channel attention modules are introduced to aggregate information across different scales. Finally, the effectiveness of ESTN on five public datasets and a local attribution map (LAM) are analyzed.
Experimental results demonstrate that the proposed ESTN achieves higher average PSNR, surpassing SRCNN, ELAN-light, SwinIR-light, and SMFANER+ models by 2.17dB, 0.13dB, 0.12dB, and 0.1dB, respectively, with LAM further confirming its larger receptive field. ESTN delivers improved quality of SR images. The source code can be found at https://github.com/huangyuming2021/ESTN.
\end{abstract}


\section{Introduction}
\IEEEPARstart{I}{mage} super-resolution (SR) reconstruction represents a fundamental challenge within the image processing domain, aiming to produce images with high spatial resolution and fine details\cite{ref1,ref2,ref3,ref4}. The image SR restores high-frequency details lost in low-resolution (LR) images. Image SR reconstruction has been extensively applied in fields such as remote sensing\cite{ref1}, infrared\cite{ref2,ref3}, and medical imaging\cite{ref4}.

Images sustain quality degradation during transmission, leading to information loss. LR images degraded from high-resolution (HR) images suffer from edge blurring caused by downsampling. The image SR reconstruction technique restores high-resolution image details from the LR image. 

The three approaches to SR reconstruction are interpolation-driven, model-driven, and data-driven methods. The interpolation-based method \cite{ref5} has been widely used for its simplicity and efficiency. However, the algorithm's reconstruction results have jaggedness and blurring issues, significantly degrading the quality of SR images. The model-driven\cite{ref6,ref7} methods employ prior image knowledge to recover detailed information but its high computational complexity imposes limitations in engineering applications. With the advancements in parallel computing technology, data-driven methods have been extensively studied, particularly image SR reconstruction to address image degradation. In recent years, deep learning--based image SR reconstruction techniques have advanced significantly by learning the mappings from LR images to HR images using large-scale paired datasets. For instance, Dong \textit{et al.}\cite{ref8} introduced an SR convolutional neural network (SRCNN) model containing only three layers for image SR reconstruction. The SRCNN model employs CNNs for image SR reconstruction, yielding better reconstruction results than interpolation- and model-driven approaches. Subsequent research enhanced the feature representation of the network model by increasing the network depth. For example, Simonyan \textit{et al.}\cite{ref9} proposed a 19-layer VGG network, and He \textit{et al.}\cite{ref10} developed a 152-layer ResNet using residual learning to mitigate network gradient vanishing and exploding issues. Furthermore, Ledig \textit{et al.}\cite{ref11} proposed an SRGAN model by integrating a residual generator with a discriminator network. Chen et al. developed MICU\cite{MICU}, which applied a U-Net-like\cite{DNNAM} network for image SR reconstruction, yielding excellent reconstruction results. The model incorporates down-channel and up-channel branches for multilevel feature compression and input feature recovery, respectively. Advanced neural network architectures, such as residual connection\cite{ref11}, and dense connectivity\cite{Wang-Esrgan, Zhang-FMEDC-MMEL} have been adopted to enhance SR reconstruction performance. Some other SR reconstruction networks apply attention mechanisms\cite{Vaswani-Transformer,Zhang-Rcan} within the CNN frameworks, achieving excellent reconstruction performance.

Existing data-driven approaches employ a convolutional structure. Although they significantly outperform traditional model-driven techniques in image reconstruction, they encounter two major problems. First, the image-convolution kernel interaction is content-independent; thus, applying the same kernel across diverse image regions may generate suboptimal results. Second, the convolution is limited in modeling long-range dependencies \cite{ref12}, often requiring deeper network layers to expand the receptive field, leading to increased computational overhead. To address this problem, Fang \textit{et al.}\cite{ref13} proposed a hybrid CNN-Transformer approach that aggregates local and global features of an image. Li \textit{et al.}\cite{Li-CFIN} proposed CFIN for lightweight SR model, integrating CNN and Transformer to balance the computational overhead and model performance. However, this method overlooks the spatial-channel feature interaction. 

The Transformer model \cite{Vaswani-Transformer,ref17,ref18} processes global information, thus showing significant potential in computer vision \cite{ref22,ref23,ref24,ref25,ref26}, target detection\cite{ref19}, target classification\cite{ref20}, and video classification\cite{ref21}. For instance, the vision transformer (ViT) \cite{ref20} leverages the Transformer architecture to capture long-range dependencies among non-overlapping image blocks, achieving superior classification performance. 

The Transformer’s larger receptive field enables greater performance in image processing than CNN-based networks do, making it effective for image SR reconstruction\cite{Li-MAST}. For instance, Chai et al. proposed CvTrans\cite{CvTrans}, a Transformer for stereoscopic omnidirectional image SR, which Transformers with dynamic convolutions to adaptively select content- and weight-aware kernels for patch-wise feature extraction. The Transformer\cite{ref20} has emerged as a promising alternative to CNN models. However, ordinary attention mechanisms \cite{CHEN-ImageInpainting}, with quadratic complexity of input length, are inefficient for HR visual tasks. To improve the computational efficiency of ViT, the Swin Transformer\cite{ref27} introduces shifted window self-attention, which lowers computational effort and enables information exchange across neighboring windows. 

\begin{figure}[!t]
\centering
\includegraphics[width=3.5in]{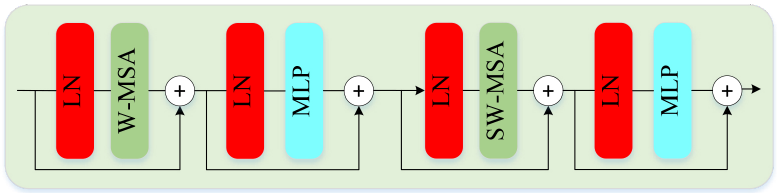}
\caption{Swin Transformer module.}
\label{fig_1}
\end{figure}

 As illustrated in Fig. \ref{fig_1}, the Swin Transformer block successively alternates between window multi-head self-attention (W-MSA) and shifted window multi-head self-attention (SW-MSA) modeling to capture local texture information of the image. The MLPs in Fig. \ref{fig_1}
 comprise two feed-forward layers with a GELU\cite{ref34} activation in between the layers for enhanced feature transformations. Layer normalization (LN)\cite{ref35} precedes MSA and MLP modules, each followed by a residual connection. The Swin Transformer’s localized attention mechanism enables efficient processing of large-scale images. The SwinIR\cite{ref12} model applies the Swin Transformer to SR reconstruction tasks, achieving strong performance metrics and computational efficiency. Chai et al. introduced TCCL-Net\cite{TCCL-Net}, which employs Swin Transformer and residual convolution blocks to extract heterogeneous features for omnidirectional image SR. Chen et al. proposed HAT\cite{ref18}, which integrates the channel attention block with the W-MSA module to form a hybrid attention block (HAB). HAB enhances Swin Transformer performance by increasing activated pixels. However, the Transformer's fixed window size limits its capacity to process objects at varying scales\cite{Choi-N-gram}. Therefore, a multiscale window mechanism is integrated into the Swin Transformer block, improving its multiscale learning capability. 

Recently, the MLP model with channel and spatial information interactions has gained attention for its simple network architecture and effective information exchange. For instance, Tolstikhin \textit{et al.} introduced the MLP-Mixer model\cite{Mlp-mixer} that expands the receptive field of the model through inter-channel and inter-space MLP operations on feature tensors. Liu \textit{et al.}\cite{Liu-Mlps} introduced a spatial gating unit (SGU) in MLP within the gated multilayer perceptron (gMLP) model, enabling interaction across different channels and spatial locations to enhance nonlinear mapping. Building on gMLP’s global receptive field, Tu \textit{et al.} developed a MAXIM\cite{Maxim} model with linear computational complexity. 

Inspired by\cite{ref13}, this study proposes an image SR reconstruction network based on an enhanced Swin Transformer, aiming to improve its receptive field The Transformer alternatively aggregates between local and global features. During local feature aggregation, a shifted convolution structure is introduced to extract local spatial features and facilitate spatial-channel feature interaction. During global feature aggregation, we employ a global sparse perception module based on gMLP\cite{Mlp-mixer} and a multiscale attention perception module to expand the receptive field. A low-parametric residual channel attention block (LRCAB) is designed to select channels efficiently with the limited number of parameters. Finally, the local attribution map (LAM)\cite{Gu-LAM} is employed to analyze the receptive field size of the proposed Enhanced Swin Transformer Network (ESTN).

The main contributions of this study are as follows: 

1) This study introduces an ESTN that enhances the interaction of the spatial and channel features of the model. This module aggregates local and global features through alternating structures and extracts the global feature, yielding a large receptive field to improve the nonlinear mapping of the network.

2) The proposed model addresses the high computational burden of the Swin Transformer's single window by incorporating a window multiscale attention mechanism to construct a more flexible spatial long-range feature interaction. Unlike the fixed-size window, the proposed model's small window reduces the amount of window attention computation (which scales quadratically with the length of the global window attention mechanism), while the large-scale window enhances the model's receptive field. Therefore, the multiscale window attention mechanism effectively balances the receptive field and computational complexity. Furthermore, it mitigates the limited long-range modeling capability caused by fixed-window size.

3) The feed-forward network (FFN) increases the number of feature channels through a linear layer, leading to inter-channel redundancy that limits feature expressiveness. Therefore, this study integrates a low-parameter residual channel attention module into the conventional FFN. This approach enhances channel-wise attention with minimal additional network parameters, effectively addressing the channel redundancy problem of FFN.

The remainder of the paper is organized as follows: Section II presents the proposed method, Section III details the experimental results, and Section IV concludes the findings of this study. 

\section{Proposed Method} 
This study introduces ESTN to enhance global perceptivity and prevent an excessive number of parameters. The model alternately aggregates local and global features for image SR. Furthermore, to assess the impact of the proposed ESTN network on the receptive field, LAM is utilized to visualize the receptive field of the reconstructed network.

\subsection{Network Architecture}

\begin{figure*}[!t]
\centering
\includegraphics[width=7.2in]{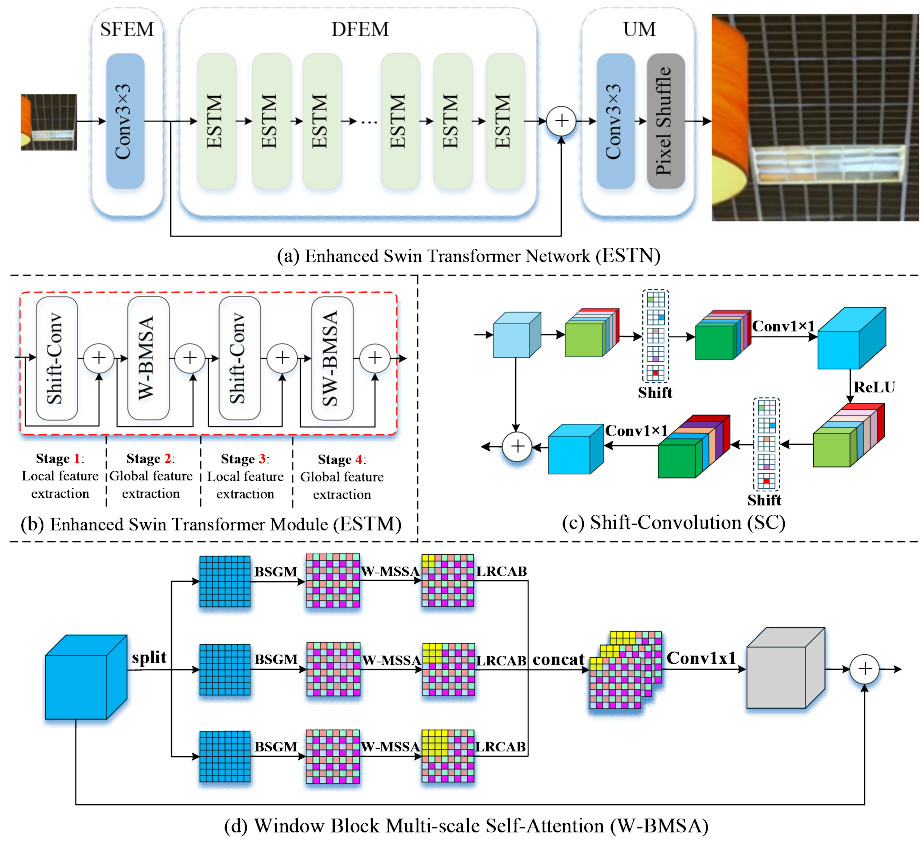}
\caption{Overview of the Enhanced Swin Transformer image SR reconstruction network. (a) Enhanced Swin Transformer Network (ESTN). (b) Enhanced Swin Transformer Module (ESTM). (c) Shift-Convolution (SC). (d) Window Block Multi-scale Self-Attention (W-BMSA).}
\label{ESTNoverview} 
\end{figure*}
The proposed SR reconstruction network, ESTN (Fig. \ref{ESTNoverview}), comprises the shallow feature extraction module (SFEM), the deep feature extraction module (DFEM), and the up-sampling module (UM). The SFEM employs a 3×3 convolution to extract shallow features. The DFEM includes multiple ESTM blocks. First, two shift-convolutions (SCs)\cite{ref36} extract local features, enhancing texture reconstruction. Second, global feature extraction integrates the block sparse global-awareness module (BSGM), window multiscale self-attention (W-MSSA), and low-parametric residual channel attention block (LRCAB) as modules. Finally, local and global features are extracted alternatively. Notably, the key innovation lies in the fourth stage, where the W-MSSA incorporates a shift operation. This shifted window multiscale self-attention (SW-MSSA) enables effective inter-window information interaction. The UM, comprising a 3×3 convolution followed by a pixel shuffle\cite{ref37}, generates SR images by enlarging them to the desired scale.

\subsubsection{Shallow and Deep Feature Extraction}
\ 

Given an LR image ${\boldsymbol{\cal I}_L} \in {\mathbb{R}^{3 \times H \times W}}$, shallow features are extracted using a convolution with a spatial resolution of 3 × 3, where each slice convolution operation is defined as:

\begin{equation}
	\label{deqn_ex1a4}
	{\boldsymbol{\cal F}_0}(c,:,:) = \boldsymbol{\cal W}_0^{3 \times 3}(c,:,:,:) * {\boldsymbol{\cal I}_L},
\end{equation}	
where  $\boldsymbol{\cal W}_0^{3 \times 3}(c,:,:,:)$ denotes the $c$-th $(c = 1,2, \cdots ,C)$ convolutional kernel in convolutional kernel set $\boldsymbol{\cal W}_0^{3 \times 3} \in {\mathbb{R}^{C \times 3 \times 3 \times 3}}$  with a spatial resolution of 3 × 3;  ${\boldsymbol{\cal F}_0} \in {\mathbb{R}^{C \times H \times W}}$  is a shallow feature; ${\boldsymbol{\cal F}_0}(c,:,:)$ denotes the $c$-th level slice of the convolution result; $c$ denotes the number of channels for intermediate features; $ * $ denotes the convolution operator. For simplicity of expression, subsequent convolutions only express the relationship between the convolution kernel, the convolved tensor, and the convolution result analogously to ${\boldsymbol{\cal F}_0} = \boldsymbol{\cal W}_0^{3 \times 3} * {\boldsymbol{\cal I}_L}$. 

\begin{equation}
\label{deqn_ex1a5}
{\boldsymbol{{\cal F}}_D} = {F_D}\left( {{\boldsymbol{\cal F}_0}} \right),
\end{equation}	
where ${F_D}$ denotes DFEM, and ${\boldsymbol{\cal F}_D} \in {\mathbb{R}^{C \times H \times W}}$ signifies deep features extracted by DEFM.

\begin{equation}
\label{deqn_ex1a6}
\left\{ \begin{array}{l}
{\boldsymbol{\cal F}_i} = {F_{{E_i}}}\left( {{\boldsymbol{\cal F}_{i - 1}}} \right),i = 1,2,...,I - 1,\\
{\boldsymbol{\cal F}_D} = {F_{{E_i}}}\left( {{\boldsymbol{\cal F}_{i - 1}}} \right),i = I,
\end{array} \right.
\end{equation}	
where ${F_{{E_i}}},i = 1,...,I$ represents the ESTM, and ${\boldsymbol{\cal F}_i},i = 1,...,I - 1$ indicates the $i$-th ESTM output feature.

\subsubsection{Up-Sampling Module}
\ 

The SR image can be recovered by summing the shallow feature ${\boldsymbol{\cal F}_0}$ and deep feature ${\boldsymbol{\cal F}_D}$, followed by a 3×3 convolution and pixel shuffle. 

\begin{equation}
\label{deqn_ex1a7}
{\boldsymbol{\cal I}_S} = {F_P}\left( {\boldsymbol{\cal W}_1^{3 \times 3} * \left( {{\boldsymbol{\cal F}_D} + {\boldsymbol{\cal F}_0}} \right)} \right),
\end{equation}	
where ${F_P}$ represents the pixel shuffle \cite{ref37} operation, ${\boldsymbol{\cal I}_S} \in {\mathbb{R}^{3 \times aH \times aW}}$ denotes the SR image, with $a$ referring to the multiplication scale, and $\boldsymbol{\cal W}_1^{3 \times 3} \in {\mathbb{R}^{3 \times C \times  3 \times 3}}$ signifies a convolutional kernel with a spatial resolution of 3×3.

\subsubsection{Loss Function}
\ 

We employ the Adam \cite{ref38} optimizer to optimize the ESTN parameters by minimizing the ${L_1}$ loss:

\begin{equation}
\label{deqn_ex1a8}
L = \frac{1}{N}\sum\limits_{n = 1}^N {{{\left\| {{\boldsymbol{\cal I}_{S,n}} - {\boldsymbol{\cal I}_{H,n}}} \right\|}_1}} ,
\end{equation}	
where ${\boldsymbol{\cal I}_{S,n}}$ and ${\boldsymbol{\cal I}_{H,n}}$ denote the $n$-th $(n = 1,2,...,N)$ SR and HR images within the batch, respectively, and$N$ refers to the number of batches.

\subsection{Enhanced Swin Transformer Model}
Existing Swin Transformer-based image SR reconstruction networks employ small attention window sizes, which restrict the modeling of long-range dependencies and degrade the quality of the recovered HR images. To address this problem, we integrate the BSGM into the Swin Transformer. Additionally, the MSA of the Swin Transformer is substituted with the MSSA to better capture multiscale information. Compared with the Swin Transformer module (Fig. \ref{fig_1}), ESTM (Fig. \ref{ESTNoverview}(b)) aggregates local and global features through alternating structures and extracts the global feature, yielding a large receptive field to improve the nonlinear mapping of the network. The signal flow diagram of each stage in the proposed ESTM is detailed below. 

\subsubsection{Stage 1: Local Feature Extraction Stage}
\ 

\begin{figure*}[!t]
\centering
\includegraphics[width=7.1in]{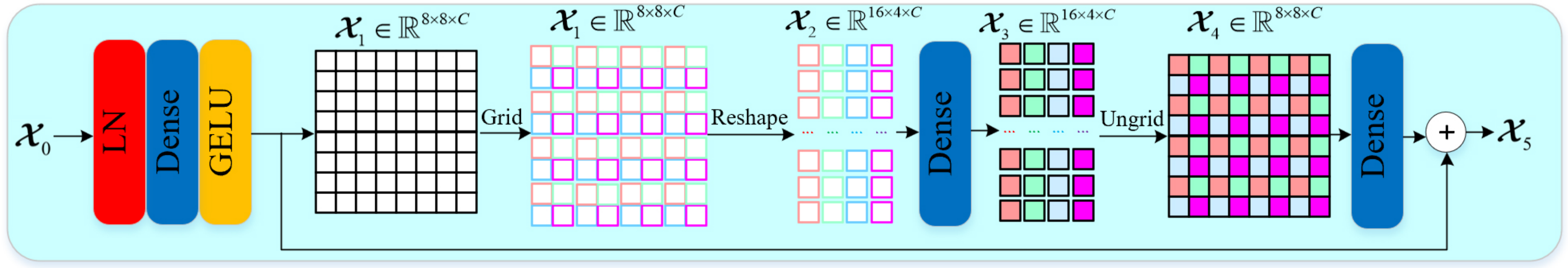}
\caption{Overview of the BSGM module.}
\label{BSGM}
\end{figure*}

\begin{figure}[!t]
    \centering
    \subfigure[]{
        \centering
        \includegraphics[width=3.5in]{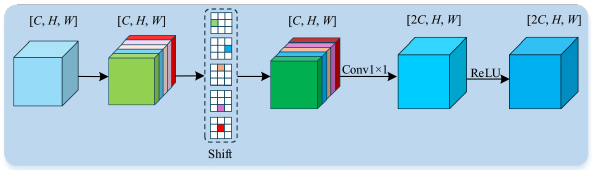}
    }
    \vfil
    \subfigure[]{
        \centering
        \includegraphics[width=3.5in]{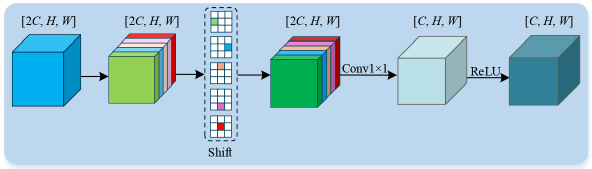}
    }
    \caption{Shift convolution structure. (a) Channel-expanded shift convolution. (b) Channel-compressed shift convolution.}
\label{fig_4}    
\end{figure}

Fig. \ref{fig_4}
 illustrates the first stage of localized feature aggregation from Fig. \ref{ESTNoverview}
. The features undergo SC and 1×1 convolution to extract local features and increase the channel dimension, respectively (Fig. \ref{fig_4}(a)).

\begin{equation}
\label{deqn_ex1a9}
\boldsymbol{\cal F}_{i,{e_0}}^{} = \sigma \left( {\boldsymbol{\cal W}_{i,{e_0}}^{1 \times 1} * {D_C}\left( {\boldsymbol{\cal W}_{i,{s_0}}^{},{\boldsymbol{\cal F}_{i - 1}}} \right)} \right),
\end{equation}  
where $\boldsymbol{\cal W}_{i,{s_0}}^{} \in {\mathbb{R}^{C \times 3 \times 3}}$ denotes a 3D tensor, signifying the SC kernel that stacks five groups of convolution kernels along the channel (Fig. \ref{fig_4}(a)); ${D_C}$ represents the channel-wise convolution operator; $\boldsymbol{\cal W}_{i,{e_0}}^{1 \times 1} \in {\mathbb{R}^{2C \times C \times 1 \times 1}}$ indicates the 1×1 convolution kernel for channel dimension expansion; $\sigma$ corresponds to the ReLU \cite{ref39} activation function; $\boldsymbol{\cal F}_{i,{e_0}}^{} \in {\mathbb{R}^{2C \times H \times W}}$ refers to the feature after channel dimension expansion.

The feature $\boldsymbol{\cal F}_{i,{e_0}}^{}$ undergoes SC and channel dimension compression through a 1×1 convolution kernel (Fig. \ref{fig_4}(b)) to match the channel dimension of the input feature ${\boldsymbol{\cal F}_{i - 1}}$.

\begin{equation}
\label{deqn_ex1a10}
{\boldsymbol{\cal F}_{i,{c_0}}} = \boldsymbol{\cal W}_{i,{c_0}}^{1 \times 1} * {D_C}\left( {\boldsymbol{\cal W}_{i,{s_1}}^{},\boldsymbol{\cal F}_{i,{e_0}}^{}} \right),
\end{equation}
where $\boldsymbol{\cal W}_{i,{s_1}}^{} \in {\mathbb{R}^{2C \times 3 \times 3}}$ implies a 3D tensor of the SC kernel that moves the features in space, $\boldsymbol{\cal W}_{i,{c_0}}^{1 \times 1} \in {\mathbb{R}^{C \times 2C \times 1 \times 1}}$ refers to the 1×1 convolution kernel for channel dimension compression, and ${\boldsymbol{\cal F}_{i,{c_0}}} \in {\mathbb{R}^{C \times H \times W}}$ represents the feature after channel dimension compression.

The local feature ${\boldsymbol{\cal F}_{i,{o_1}}}$ is obtained through residual connections between feature ${\boldsymbol{\cal F}_{i - 1}}$ and feature ${\boldsymbol{\cal F}_{i,{c_0}}}$ after channel dimension compression.

\begin{equation}
\label{deqn_ex1a11}
{\boldsymbol{\cal F}_{i,{o_1}}} = {\boldsymbol{\cal F}_{i,{c_0}}} + {\boldsymbol{\cal F}_{i - 1}}.
\end{equation}

\subsubsection{Stage 2: Global Feature Extraction Stage}
\

{\bf{Block sparse global-awareness module in stage 2}}

This study employs the BSGM to build sparse global awareness of features.

\begin{equation}
\label{deqn_ex1a12}
{\boldsymbol{\cal F}_{i,{B_0}}} = {F_{i,{B_0}}}\left( {{\boldsymbol{\cal F}_{i,{o_1}}}} \right),
\end{equation}
where ${F_{i,{B_0}}}$ represents the BSGM of stage 2 within the $i$-th ESTM.

The detail of ${F_{i,{B_0}}}$ is as follows. Assume that the input tensor of BSGM is ${\boldsymbol{\cal X}_0} \in {\mathbb{R}^{8 \times 8 \times C}}$, which is shown in Fig. \ref{BSGM} (the size of $\boldsymbol{\cal X}_0$ is just for explanation), where ${\boldsymbol{\cal X}_0}$ undergoes layer normalization, channel dimension feature mapping, and the GELU \cite{ref34} activation function, yielding ${\boldsymbol{\cal X}_1} \in {\mathbb{R}^{8 \times 8 \times C}}$:

\begin{equation}
\label{deqn_ex1a13}
{\boldsymbol{\cal X}_1} = g\left( {D\left( {LN\left( {{\boldsymbol{\cal X}_0}} \right)} \right)} \right),
\end{equation}
where $D$ denotes the fully connected feature mapping layer impacts on the last axis of the processed tensor. 

Feature $\boldsymbol{\cal X}_1$ undergoes a spatial mapping to yield $\boldsymbol{\cal X}_2$ :

\begin{equation}
\label{deqn_ex1a14}
{\boldsymbol{\cal X}_2} = Reshape\left( {Grid\left( {{\boldsymbol{\cal X}_1}} \right)} \right),
\end{equation}    
where $Grid$ denotes tensor partitioning (for illustration, Fig. \ref{BSGM} uses a 2×2 window size to represent the information reorganization process; in the proposed network architecture, a 4×4 window size is employed); $Reshape$ represents the reorganization of the tensor's spatial arrangement.

After that, a fully connected feature mapping layer is introduced to obtain global information, that is
\begin{equation}
\label{deqn_ex1a15}
{\boldsymbol{\cal X}_3}{\rm{ = }}D_f\left( {{\boldsymbol{\cal X}_2}} \right),
\end{equation}
where $D_f$ is a fully connected feature mapping layer impacts on the first axis of the processed tensor.

Then, a reshape operator is required, that is 

\begin{equation}
\label{deqn_ex1a16}
{\boldsymbol{\cal X}_4}{\rm{ = }}Ungrid\left( {{\boldsymbol{\cal X}_3}} \right),
\end{equation}
where $Ungrid$ denotes the recovery tensor having the original shape. 

A fully connected feature mapping is applied along the channel direction to tensor ${\boldsymbol{\cal X}_4}$. After full connection, tensor ${\boldsymbol{\cal X}_4}$ is residually connected to tensor ${\boldsymbol{\cal X}_1}$ to produce tensor ${\boldsymbol{\cal X}_5}$.

\begin{equation}
\label{deqn_ex1a17}
{\boldsymbol{\cal X}_5} = D\left( {{\boldsymbol{\cal X}_4}} \right) + {\boldsymbol{\cal X}_1}{\rm{.}}
\end{equation}

{\bf{W-MSSA module in stage 2}}

This study introduces W-MSSA to enable the learning of multiscale information.
\begin{figure}[!t]
\centering
\includegraphics[width=3.5in]{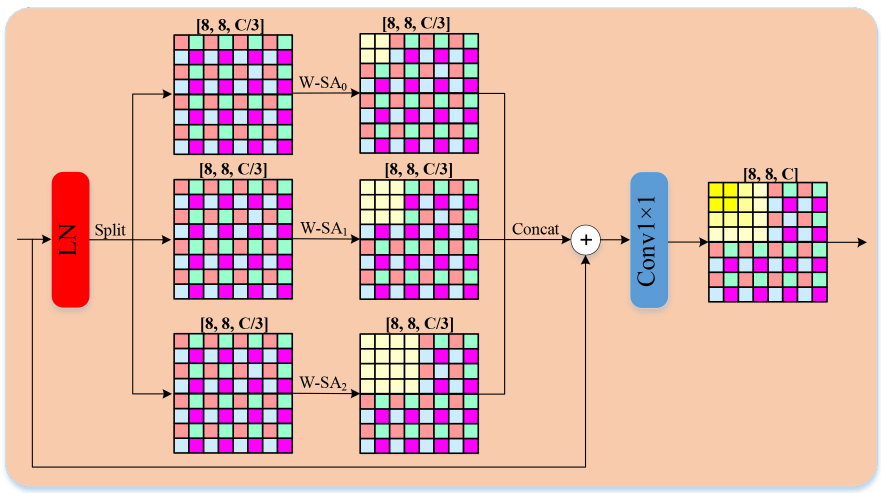}
\caption{Overview of the W-MSSA module.}
\label{W-MSSA}
\end{figure}

\begin{equation}
\label{deqn_ex1a18}
{\boldsymbol{\cal F}_{i,W}} = {F_{i,{W_{}}}}\left( {{\boldsymbol{\cal F}_{i,{B_0}}}} \right),
\end{equation}    
where ${F_{i,{W_{}}}}$ denotes the W-MSSA module for stage 2 of the $i$-th ESTM.

\begin{figure}[!t]
\centering
\includegraphics[width=3.5in]{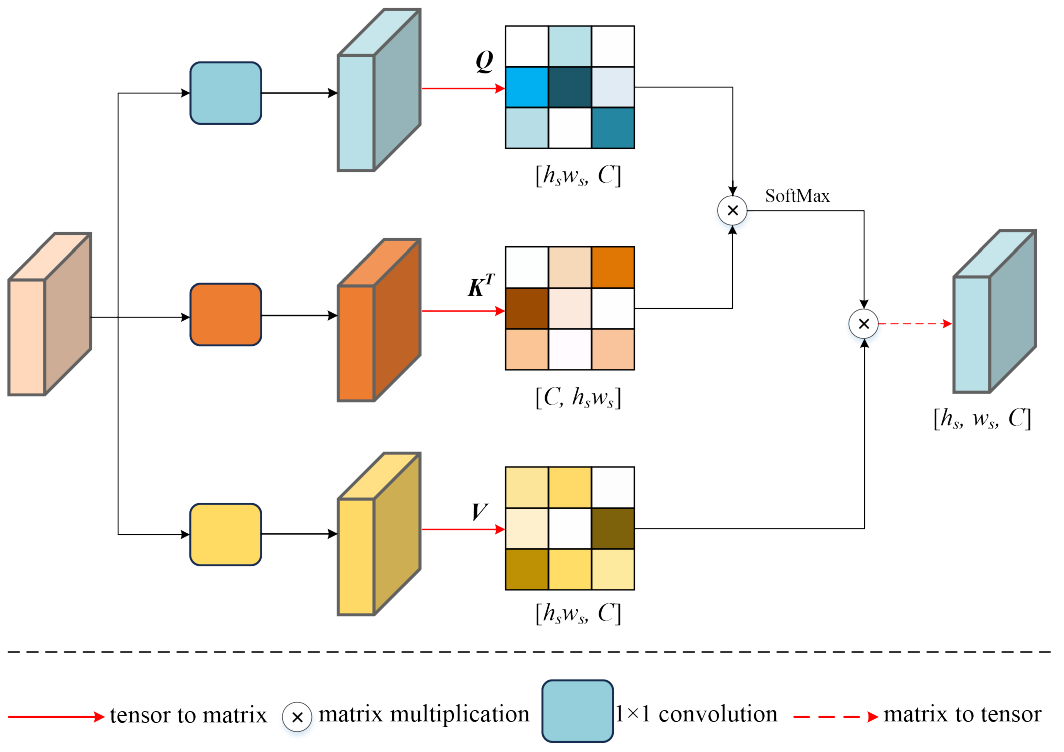}
\caption{Overview of the self-attention mapping.}
\label{SA}
\end{figure}

The MSSA computes the multiscale self-attention after the BSGM establishes sparse global feature awareness. As illustrated in Fig. \ref{W-MSSA}, the tensor is first split into three equal parts along the channel dimension. Attention matrices are then computed for each of the three scales using the W-SA$s$ ($s$=0,1,2) module to handle objects at each scale (self-attention ranges are marked in yellow in the figure). Self-attention matrix acquisition is illustrated in Fig. \ref{SA}, where the query matrix $\boldsymbol{Q}$, key matrix $\boldsymbol{{K^T}}$, and value matrix $\boldsymbol{V}$ are derived through 1×1 convolutions. Reflective padding is applied at the image boundaries to ensure that the image size is an integer multiple of each window size. The self-attention is calculated as follows.

\begin{equation}
\label{deqn_ex1a19}
Attention\left( {\boldsymbol{Q},\boldsymbol{K},\boldsymbol{V}} \right) = SoftMax\left( {\frac{{\boldsymbol{Q}{\boldsymbol{K}^T}}}{{\sqrt {{h_s}{w_s}} }}} \right)\boldsymbol{V},
\end{equation}  
where $SoftMax\left( \boldsymbol{X} \right)$ denotes an operator that applies the exponential function to each element of the matrix $\boldsymbol{X}$ and then normalizes each row independently so that the sum of each row equals one; $[{h_s},{w_s}]\left( {s = 0,1,2} \right)$ signifies the size of the local window. After the calculation of self-attention, a reshape operator is required, which is shown in Fig. \ref{SA}

{\bf{Low-parametric residual channel attention module in stage 2}}

\begin{figure*}[!t]
\centering
\includegraphics[width=7in]{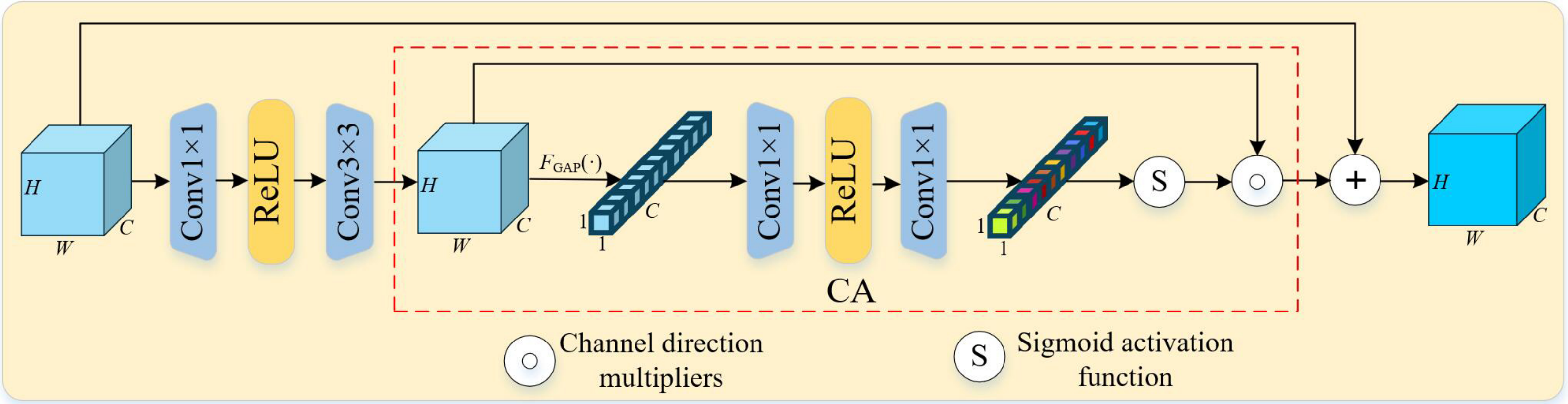}
\caption{Overview of the LRCAB block.}
\label{LRCAB}
\end{figure*}
The attention mechanism is widely implemented in image processing for its superior performance. Since each channel feature contributes differently to the SR reconstruction, this study incorporates channel attention to focus on the channels of the features selectively.

As illustrated in Fig. \ref{LRCAB}, the dimensionality of the input feature ${\boldsymbol{\cal F}_i} \in {\mathbb{R}^{C \times H \times W}}$ is expanded via 1×1 convolution. Increasing the feature dimensionality enhances the capturing of richer features, such as textures across various directions and frequencies. Subsequently, a 3×3 convolution is employed to adapt and recover the features back to the original input dimensionality. Finally, the feature channels are selected via the channel attention module.

\begin{equation}
\label{deqn_ex1a20}
{\boldsymbol{\cal F}_{i,C}} = {F_C}\left( {\boldsymbol{\cal W}_{i,{c_1}}^{3 \times 3} * \sigma \left( {\boldsymbol{\cal W}_{i,{e_1}}^{1 \times 1} * {\boldsymbol{\cal F}_{i,W}}} \right)} \right) + {\boldsymbol{\cal F}_{i,W}},
\end{equation} 
where $\boldsymbol{\cal W}_{i,{e_1}}^{1 \times 1} \in {\mathbb{R}^{2C \times C \times 1 \times 1}}$  denotes the 1×1 convolution kernel for channel dimension expansion; $\boldsymbol{\cal W}_{i,{c_1}}^{3 \times 3} \in {\mathbb{R}^{C \times 2C \times 3 \times 3}}$  denotes the 3×3 convolution kernel for channel dimension compression.

\begin{equation}
\label{deqn_ex1a21}
{F_C}\left( \boldsymbol{\cal X} \right) = Sigmoid\left( {\boldsymbol{\cal W}_{i,{e_2}}^{1 \times 1} * \sigma \left( {\boldsymbol{\cal W}_{i,{c_2}}^{1 \times 1} * {F_{{\rm{GAP}}}}\left( \boldsymbol{\cal X} \right)} \right)} \right) \circ \boldsymbol{\cal X},
\end{equation}
where  $\boldsymbol{\cal W}_{i,{e_2}}^{1 \times 1} \in {\mathbb{R}^{2C  \times C  \times 1 \times 1}}$ denotes the 1×1 convolution kernel for channel dimension expansion;  $\boldsymbol{\cal W}_{i,{c_2}}^{1 \times 1} \in {\mathbb{R}^{C  \times 2C  \times 1 \times 1}}$ denotes the 1×1 convolution kernel for channel dimension compression; ${F_{{\rm{GAP}}}}$  denotes the 2D global average pooling function; $Sigmoid$ is the activation function;   $ \circ $ denotes the channel direction multiplication symbol.

The tensor ${{\cal F}_{i,{o_1}}}$ is residually connected to the tensor ${{\cal F}_{i,C}}$ to obtain the tensor ${{\cal F}_{i,{o_2}}}$.
\begin{equation}
\label{deqn_ex1a233}
{\boldsymbol{\cal F}_{i,{o_2}}} = {\boldsymbol{\cal F}_{i,{o_1}}} + {\boldsymbol{\cal F}_{i,C}}.
\end{equation}

\subsubsection{Stage 3: Local Feature Extraction Stage}
\ 

Eqs. \eqref{deqn_ex1a22}--\eqref{deqn_ex1a23} present the mathematical expressions for the third stage of localized feature extraction in Fig. \ref{ESTNoverview}, which is similar to stage 1.

\begin{equation}
\label{deqn_ex1a22}
\boldsymbol{\cal F}_{i,{e_3}}^{} = \sigma \left( {\boldsymbol{\cal W}_{i,{e_3}}^{1 \times 1} * {D_C}\left( {\boldsymbol{\cal W}_{i,{s_0}}^{},{\boldsymbol{\cal F}_{i,C}}} \right)} \right),
\end{equation}
    
\begin{equation}
\label{deqn_ex1a23}
{\boldsymbol{\cal F}_{i,{o_3}}} = \boldsymbol{\cal W}_{i,{c_3}}^{1 \times 1} * {D_C}\left( {\boldsymbol{\cal W}_{i,{s_1}}^{},\boldsymbol{\cal F}_{i,{e_3}}^{}} \right) + {\boldsymbol{\cal F}_{i,o_2}},
\end{equation}
where $\boldsymbol{\cal W}_{i,{e_3}}^{1 \times 1} \in {\mathbb{R}^{2C \times C \times  1 \times 1}}$  denotes the 1×1 convolution kernel for channel dimension expansion; $\boldsymbol{\cal W}_{i,{c_3}}^{1 \times 1} \in {\mathbb{R}^{C  \times 2C  \times 1 \times 1}}$ denotes the 1×1 convolution kernel for channel dimension compression; ${\boldsymbol{\cal F}_{i,{s_1}}}$  denotes the localized features of the shifted convolutional output of the third stage in ESTM.

\subsubsection{Stage 4: Global Feature Extraction Stage}
\ 

{\bf{BSGM in stage 4}}

Eq. \eqref{deqn_ex1a24} defines the sparse global awareness learning process at stage 4 of the ESTM in Fig. \ref{ESTNoverview}, similar to BSGM in stage 2.

\begin{equation}
\label{deqn_ex1a24}
{\boldsymbol{\cal F}_{i,{B_1}}} = {F_{i,{B_1}}}\left( {{\boldsymbol{\cal F}_{i,{o_3}}}} \right),
\end{equation}    
where ${F_{i,{B_1}}}$ denotes the BSGM of stage 4 in the $i$-th ESTM.

{\bf{SW-MSSA module in stage 4}}

\begin{figure}[!t]
\centering
\includegraphics[width=3.5in]{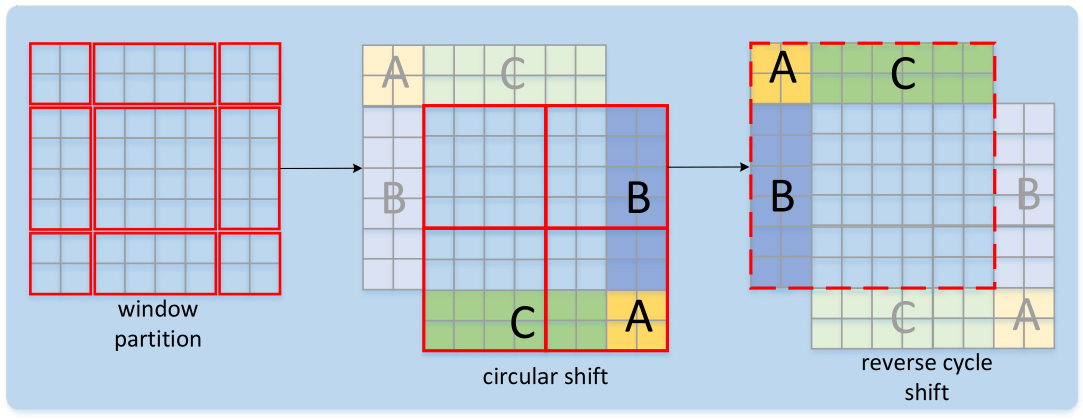}
\caption{Shift window operator.}
\label{fig_8}
\end{figure}
Fig. \ref{fig_8} illustrates that SW-MSSA adds cyclic and inverse cyclic shift operations compared to W-MSSA. The circular shift distance is half of the current window size. The SW-MSSA is computed using Eq. \eqref{deqn_ex1a25}.

\begin{equation}
\label{deqn_ex1a25}
{\boldsymbol{\cal F}_{i,SW}} = {F_{i,SW}}\left( {{\boldsymbol{\cal F}_{i,{B_1}}}} \right),
\end{equation}
where ${F_{i,SW}}$ denotes the SW-MSSA module in stage 4 of the $i$-th ESTM.

{\bf{Low-parameter residual channel attention module in stage 4}}

Similar to LRCAB in stage 2, channel attention is computed to reassign channel weights as follows:

\begin{equation}
\label{deqn_ex1a26}
\left\{ \begin{array}{l}
{\boldsymbol{\cal F}_i} = {F_{i,{L_1}}}\left( {{\boldsymbol{\cal F}_{i,SW}}} \right) + {\boldsymbol{\cal F}_{i,{o_3}}},i = 1,2,...,I - 1,\\
{\boldsymbol{\cal F}_D} = {F_{i,{L_1}}}\left( {{\boldsymbol{\cal F}_{i,SW}}} \right) + {\boldsymbol{\cal F}_{i,{o_3}}},i = I,
\end{array} \right.
\end{equation}  
where ${F_{i,{L_1}}}$ signifies the LRCAB in stage 4 of the $i$-th ESTM.

The ESTN is summarized in Algorithm 1.

\begin{algorithm}[H]
  \SetAlgoLined
  \label{alg:algorithm1}
  \KwIn{LR images ${\boldsymbol{\cal I}_L}$, number of deep feature extraction module $I$}
  \KwOut{SR images ${\boldsymbol{\cal I}_S}$}

  Shallow features ${\boldsymbol{\cal F}_0}$ are extracted by 3$\times$3 convolution of the LR image ${\boldsymbol{\cal I}_L}$\;
  
  \For{$i\leftarrow 1$ \KwTo $I$}{

   Extract local features via via Eqs. \eqref{deqn_ex1a9}--\eqref{deqn_ex1a11};\

   Extract global features via BSGM, using Eq. \eqref{deqn_ex1a12};\
    
    Extract global features via W-MSSA, using Eq. \eqref{deqn_ex1a18};\

    Extract global features via LRCAB, using Eqs. \eqref{deqn_ex1a20}--\eqref{deqn_ex1a233};\

    Extract local features via SC, using Eqs. \eqref{deqn_ex1a22}--\eqref{deqn_ex1a23};\
    
    Extract global features via BSGM, using Eq. \eqref{deqn_ex1a24};\
    
    Extract global features via W-MSSA, using Eq. \eqref{deqn_ex1a25};\
    
    Extract global features via LRCAB, using Eq. \eqref{deqn_ex1a26};

  }

  The SR image ${\boldsymbol{\cal {I}}_S}$ is obtained by pixel shuffle of Eq. \eqref{deqn_ex1a8} on feature upscaling

  \caption{Enhanced Swin Transformer SR Reconstruction Network}
\end{algorithm}

\subsection{Local Attribution Maps}
\begin{figure}[!t]
\centering
\includegraphics[width=3.5in]{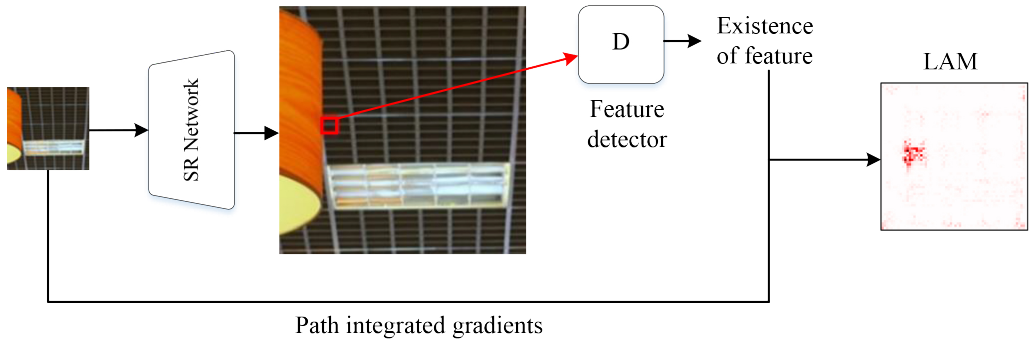}
\caption{LAM architecture: Red pixels within the LAM indicate stronger contributions to the recovery of the boxed region. }
\label{fig_9}
\end{figure}
To explore the global information modeling capability of the proposed BSGM, this study introduces LAM, which employs path integrals for gradient backpropagation to compute local features within SR from corresponding LR image pixels. Fig. \ref{fig_9} reveals that the SR reconstruction network transforms LR images into SR images. A portion of the SR image is then selected for feature extraction, and the contribution of each LR pixel to the region's features is analyzed, with denser red pixels in the LAM results signifying a higher contribution to recovering the features of the selected region. The LAM result for dimension $k(k = 0,1,2,..., K)$ is computed using the Eq. \eqref{deqn_ex1a27}.

\begin{equation}
\label{deqn_ex1a27}
{\rm{LA}}{{\rm{M}}_{F,D\left( \gamma \right)k}}: = \int_0^1 {\frac{{\partial D\left( {F\left( {\gamma \left( \alpha \right)} \right)} \right)}}{{\partial \gamma {{\left( \alpha \right)}_k}}} \times \frac{{\partial \gamma {{\left( \alpha \right)}_k}}}{{\partial \alpha }}d\alpha } ,
\end{equation}
where $F$ and $D$ signify the SR network and the local feature extractor, respectively; $\gamma \left( \alpha \right):[0,1] \to {\mathbb{R}^{H \times W}}$ implies the smoothing path function; $\gamma \left( 0 \right)$ refers to the image obtained by blurring the input image ${\boldsymbol{\cal I}_L}^\prime $; and $\gamma \left( 1 \right)$ signifies the image ${\boldsymbol{\cal I}_L}$ of the input without blurring.


\section{Experiments} 
This study compares the proposed ESTN with the state-of-the-art SR networks through ×2, ×3, and ×4 upscale single-image SR experiments on five datasets. Quantitative and qualitative results demonstrate its superior performance. Comprehensive ablation experiments validate the contribution of each component of the proposed ESTN. Finally, LAMs are adopted to visualize and analyze the receptive fields of the proposed ESTN.

\begin{table*}
\begin{center}
\caption{Quantitative Comparison of Average PSNR and SSIM with Lightweight Image SR methods on Benchmark Datasets\label{tab:table1}}
\resizebox{\textwidth}{!}{
\begin{tabular}{|
>{\columncolor[HTML]{FFFFFF}}l |
>{\columncolor[HTML]{FFFFFF}}l |
>{\columncolor[HTML]{FFFFFF}}c |
>{\columncolor[HTML]{FFFFFF}}c |
>{\columncolor[HTML]{FFFFFF}}c |
>{\columncolor[HTML]{FFFFFF}}l 
>{\columncolor[HTML]{FFFFFF}}l |
>{\columncolor[HTML]{FFFFFF}}l 
>{\columncolor[HTML]{FFFFFF}}l |
>{\columncolor[HTML]{FFFFFF}}l 
>{\columncolor[HTML]{FFFFFF}}l |
>{\columncolor[HTML]{FFFFFF}}l 
>{\columncolor[HTML]{FFFFFF}}l |
>{\columncolor[HTML]{FFFFFF}}l 
>{\columncolor[HTML]{FFFFFF}}l |}
\hline
\multicolumn{1}{|c|}{\cellcolor[HTML]{FFFFFF}}                         & \multicolumn{1}{c|}{\cellcolor[HTML]{FFFFFF}}                        & \cellcolor[HTML]{FFFFFF}                                                                      & \cellcolor[HTML]{FFFFFF}                                                                         & \cellcolor[HTML]{FFFFFF}                                                                       & \multicolumn{2}{c|}{\cellcolor[HTML]{FFFFFF}Set5}                                                               & \multicolumn{2}{c|}{\cellcolor[HTML]{FFFFFF}Set14}                                                              & \multicolumn{2}{c|}{\cellcolor[HTML]{FFFFFF}BSD100}                                                             & \multicolumn{2}{c|}{\cellcolor[HTML]{FFFFFF}Urban100}                                                           & \multicolumn{2}{c|}{\cellcolor[HTML]{FFFFFF}Manga109}                                                           \\ \cline{6-15} 
\multicolumn{1}{|c|}{\multirow{-2}{*}{\cellcolor[HTML]{FFFFFF}Method}} & \multicolumn{1}{c|}{\multirow{-2}{*}{\cellcolor[HTML]{FFFFFF}Scale}} & \multirow{-2}{*}{\cellcolor[HTML]{FFFFFF}\begin{tabular}[c]{@{}c@{}}FLOPs\\ (G)\end{tabular}} & \multirow{-2}{*}{\cellcolor[HTML]{FFFFFF}\begin{tabular}[c]{@{}c@{}}Latency\\ (ms)\end{tabular}} & \multirow{-2}{*}{\cellcolor[HTML]{FFFFFF}\begin{tabular}[c]{@{}c@{}}Params\\ (K)\end{tabular}} & \multicolumn{1}{c|}{\cellcolor[HTML]{FFFFFF}PSNR}           & \multicolumn{1}{c|}{\cellcolor[HTML]{FFFFFF}SSIM} & \multicolumn{1}{c|}{\cellcolor[HTML]{FFFFFF}PSNR}           & \multicolumn{1}{c|}{\cellcolor[HTML]{FFFFFF}SSIM} & \multicolumn{1}{c|}{\cellcolor[HTML]{FFFFFF}PSNR}           & \multicolumn{1}{c|}{\cellcolor[HTML]{FFFFFF}SSIM} & \multicolumn{1}{c|}{\cellcolor[HTML]{FFFFFF}PSNR}           & \multicolumn{1}{c|}{\cellcolor[HTML]{FFFFFF}SSIM} & \multicolumn{1}{c|}{\cellcolor[HTML]{FFFFFF}PSNR}           & \multicolumn{1}{c|}{\cellcolor[HTML]{FFFFFF}SSIM} \\ \hline
SRCNN\cite{ref8}                                                                  & ×2                                                                   & -                                                                                             & -                                                                                                & 57                                                                                             & \multicolumn{1}{l|}{\cellcolor[HTML]{FFFFFF}36.66}          & 0.9542                                            & \multicolumn{1}{l|}{\cellcolor[HTML]{FFFFFF}32.42}          & 0.9063                                            & \multicolumn{1}{l|}{\cellcolor[HTML]{FFFFFF}31.36}          & 0.8879                                            & \multicolumn{1}{l|}{\cellcolor[HTML]{FFFFFF}29.50}          & 0.8946                                            & \multicolumn{1}{l|}{\cellcolor[HTML]{FFFFFF}35.74}          & 0.9661                                            \\ \hline
CARN\cite{ref47}                                                                   & ×2                                                                   & 222.8                                                                                         & 72                                                                                               & 1592                                                                                           & \multicolumn{1}{l|}{\cellcolor[HTML]{FFFFFF}37.76}          & 0.9590                                            & \multicolumn{1}{l|}{\cellcolor[HTML]{FFFFFF}33.52}          & 0.9166                                            & \multicolumn{1}{l|}{\cellcolor[HTML]{FFFFFF}32.09}          & 0.8978                                            & \multicolumn{1}{l|}{\cellcolor[HTML]{FFFFFF}31.92}          & 0.9256                                            & \multicolumn{1}{l|}{\cellcolor[HTML]{FFFFFF}38.36}          & 0.9765                                            \\ \hline
IMDN\cite{ref48}                                                                   & ×2                                                                   & 158.8                                                                                         & 54                                                                                               & 694                                                                                            & \multicolumn{1}{l|}{\cellcolor[HTML]{FFFFFF}38.00}          & 0.9605                                            & \multicolumn{1}{l|}{\cellcolor[HTML]{FFFFFF}33.63}          & 0.9177                                            & \multicolumn{1}{l|}{\cellcolor[HTML]{FFFFFF}32.19}          & 0.8996                                            & \multicolumn{1}{l|}{\cellcolor[HTML]{FFFFFF}32.17}          & 0.9283                                            & \multicolumn{1}{l|}{\cellcolor[HTML]{FFFFFF}38.88}          & 0.9774                                            \\ \hline
LAPAR-A\cite{ref49}                                                                & ×2                                                                   & 171.0                                                                                         & 73                                                                                               & 548                                                                                            & \multicolumn{1}{l|}{\cellcolor[HTML]{FFFFFF}38.01}          & 0.9605                                            & \multicolumn{1}{l|}{\cellcolor[HTML]{FFFFFF}33.62}          & 0.9183                                            & \multicolumn{1}{l|}{\cellcolor[HTML]{FFFFFF}32.19}          & 0.8999                                            & \multicolumn{1}{l|}{\cellcolor[HTML]{FFFFFF}32.10}          & 0.9283                                            & \multicolumn{1}{l|}{\cellcolor[HTML]{FFFFFF}38.67}          & 0.9772                                            \\ \hline
ESRT\cite{ref50}                                                                   & ×2                                                                   & -                                                                                             & -                                                                                                & 677                                                                                            & \multicolumn{1}{l|}{\cellcolor[HTML]{FFFFFF}38.03}          & 0.9600                                            & \multicolumn{1}{l|}{\cellcolor[HTML]{FFFFFF}33.75}          & 0.9184                                            & \multicolumn{1}{l|}{\cellcolor[HTML]{FFFFFF}32.25}          & 0.9001                                            & \multicolumn{1}{l|}{\cellcolor[HTML]{FFFFFF}32.58}          & 0.9318                                            & \multicolumn{1}{l|}{\cellcolor[HTML]{FFFFFF}39.12}          & 0.9774                                            \\ \hline
ELAN-light\cite{ref51}                                                             & ×2                                                                   & 168.4                                                                                         & 230                                                                                              & 582                                                                                            & \multicolumn{1}{l|}{\cellcolor[HTML]{FFFFFF}38.17}          & \underline{0.9611}                                            & \multicolumn{1}{l|}{\cellcolor[HTML]{FFFFFF}\textbf{33.94}}          & \underline{0.9207}                                            & \multicolumn{1}{l|}{\cellcolor[HTML]{FFFFFF}32.30}          & \underline{0.9012}                                            & \multicolumn{1}{l|}{\cellcolor[HTML]{FFFFFF}\underline{32.76}}          & \underline{0.9340}                                            & \multicolumn{1}{l|}{\cellcolor[HTML]{FFFFFF}39.11}          & 0.9782                                            \\ \hline
SwinIR-light\cite{ref12}                                                           & ×2                                                                   & 195.6                                                                                         & 1007                                                                                             & 878                                                                                            & \multicolumn{1}{l|}{\cellcolor[HTML]{FFFFFF}38.14}          & \underline{0.9611}                                            & \multicolumn{1}{l|}{\cellcolor[HTML]{FFFFFF}33.86}          & 0.9206                                            & \multicolumn{1}{l|}{\cellcolor[HTML]{FFFFFF}\underline{32.31}}          & \underline{0.9012}                                            & \multicolumn{1}{l|}{\cellcolor[HTML]{FFFFFF}\underline{32.76}}          & \underline{0.9340}                                            & \multicolumn{1}{l|}{\cellcolor[HTML]{FFFFFF}39.12}          & \textbf{0.9783}                                            \\ \hline
DIPNet\cite{DIPNet}                                                                 & ×2                                                                   & -                                                                                             & -                                                                                                & 527                                                                                            & \multicolumn{1}{l|}{\cellcolor[HTML]{FFFFFF}37.98}          & 0.9605                                            & \multicolumn{1}{l|}{\cellcolor[HTML]{FFFFFF}33.66}          & 0.9192                                            & \multicolumn{1}{l|}{\cellcolor[HTML]{FFFFFF}32.20}          & 0.9002                                            & \multicolumn{1}{l|}{\cellcolor[HTML]{FFFFFF}32.31}          & 0.9302                                            & \multicolumn{1}{l|}{\cellcolor[HTML]{FFFFFF}38.62}          & 0.9770                                            \\ \hline
SMFANET\cite{SMFANET}                                                               & ×2                                                                   & 108.0                                                                                         & -                                                                                                & 480                                                                                            & \multicolumn{1}{l|}{\cellcolor[HTML]{FFFFFF}\underline{38.18}}          & \underline{0.9611}                                            & \multicolumn{1}{l|}{\cellcolor[HTML]{FFFFFF}33.82}          & 0.9202                                            & \multicolumn{1}{l|}{\cellcolor[HTML]{FFFFFF}32.28}          & 0.9011                                            & \multicolumn{1}{l|}{\cellcolor[HTML]{FFFFFF}32.64}          & 0.9323                                            & \multicolumn{1}{l|}{\cellcolor[HTML]{FFFFFF}\underline{39.25}}          & 0.9777                                            \\ \hline
ESTN (ours)                                                            & ×2                                                                   & 283.4                                                                                         & 732                                                                                              & 863                                                                                            & \multicolumn{1}{l|}{\cellcolor[HTML]{FFFFFF}\textbf{\begin{tabular}[c]{@{}l@{}}38.23\\ 0.13\%↑\end{tabular}}} & \textbf{\begin{tabular}[c]{@{}l@{}}0.9615\\ 0.04\%↑\end{tabular}}                                   & \multicolumn{1}{l|}{\cellcolor[HTML]{FFFFFF}\textbf{\begin{tabular}[c]{@{}l@{}}33.94\\ 0.00\%↑\end{tabular}}} & \textbf{\begin{tabular}[c]{@{}l@{}}0.9213\\ 0.07\%↑\end{tabular}}                                   & \multicolumn{1}{l|}{\cellcolor[HTML]{FFFFFF}\textbf{\begin{tabular}[c]{@{}l@{}}32.34\\ 0.09\%↑\end{tabular}}} & \textbf{\begin{tabular}[c]{@{}l@{}}0.9019\\ 0.08\%↑\end{tabular}}                                   & \multicolumn{1}{l|}{\cellcolor[HTML]{FFFFFF}\textbf{\begin{tabular}[c]{@{}l@{}}32.90\\ 0.43\%↑\end{tabular}}} & \textbf{\begin{tabular}[c]{@{}l@{}}0.9357\\ 0.18\%↑\end{tabular}}                                   & \multicolumn{1}{l|}{\cellcolor[HTML]{FFFFFF}\textbf{\begin{tabular}[c]{@{}l@{}}39.27\\ 0.05\%↑\end{tabular}}} & \textbf{\begin{tabular}[c]{@{}l@{}}0.9783\\ 0.00\%↑\end{tabular}}                                   \\ \hline
SRCNN\cite{ref8}                                                                  & ×3                                                                   & -                                                                                             & -                                                                                                & 57                                                                                             & \multicolumn{1}{l|}{\cellcolor[HTML]{FFFFFF}32.75}          & 0.9090                                            & \multicolumn{1}{l|}{\cellcolor[HTML]{FFFFFF}29.28}          & 0.8209                                            & \multicolumn{1}{l|}{\cellcolor[HTML]{FFFFFF}28.41}          & 0.7863                                            & \multicolumn{1}{l|}{\cellcolor[HTML]{FFFFFF}26.24}          & 0.7989                                            & \multicolumn{1}{l|}{\cellcolor[HTML]{FFFFFF}30.59}          & 0.9107                                            \\ \hline
CARN\cite{ref47}                                                                   & ×3                                                                   & 118.8                                                                                         & 39                                                                                               & 1592                                                                                           & \multicolumn{1}{l|}{\cellcolor[HTML]{FFFFFF}34.29}          & 0.9255                                            & \multicolumn{1}{l|}{\cellcolor[HTML]{FFFFFF}30.29}          & 0.8407                                            & \multicolumn{1}{l|}{\cellcolor[HTML]{FFFFFF}29.06}          & 0.8034                                            & \multicolumn{1}{l|}{\cellcolor[HTML]{FFFFFF}28.06}          & 0.8493                                            & \multicolumn{1}{l|}{\cellcolor[HTML]{FFFFFF}33.50}          & 0.9440                                            \\ \hline
IMDN\cite{ref48}                                                                   & ×3                                                                   & 71.5                                                                                          & 27                                                                                               & 703                                                                                            & \multicolumn{1}{l|}{\cellcolor[HTML]{FFFFFF}34.36}          & 0.9270                                            & \multicolumn{1}{l|}{\cellcolor[HTML]{FFFFFF}30.32}          & 0.8417                                            & \multicolumn{1}{l|}{\cellcolor[HTML]{FFFFFF}29.09}          & 0.8046                                            & \multicolumn{1}{l|}{\cellcolor[HTML]{FFFFFF}28.17}          & 0.8519                                            & \multicolumn{1}{l|}{\cellcolor[HTML]{FFFFFF}33.61}          & 0.9445                                            \\ \hline
LAPAR-A\cite{ref49}                                                                & ×3                                                                   & 114.0                                                                                         & 55                                                                                               & 544                                                                                            & \multicolumn{1}{l|}{\cellcolor[HTML]{FFFFFF}34.36}          & 0.9267                                            & \multicolumn{1}{l|}{\cellcolor[HTML]{FFFFFF}30.34}          & 0.8421                                            & \multicolumn{1}{l|}{\cellcolor[HTML]{FFFFFF}29.11}          & 0.8054                                            & \multicolumn{1}{l|}{\cellcolor[HTML]{FFFFFF}28.15}          & 0.8523                                            & \multicolumn{1}{l|}{\cellcolor[HTML]{FFFFFF}33.51}          & 0.9441                                            \\ \hline
ESRT\cite{ref50}                                                                   & ×3                                                                   & -                                                                                             & -                                                                                                & 770                                                                                            & \multicolumn{1}{l|}{\cellcolor[HTML]{FFFFFF}34.42}          & 0.9268                                            & \multicolumn{1}{l|}{\cellcolor[HTML]{FFFFFF}30.43}          & 0.8433                                            & \multicolumn{1}{l|}{\cellcolor[HTML]{FFFFFF}29.15}          & 0.8063                                            & \multicolumn{1}{l|}{\cellcolor[HTML]{FFFFFF}28.66}          & \underline{0.8624}                                            & \multicolumn{1}{l|}{\cellcolor[HTML]{FFFFFF}33.95}          & 0.9455                                            \\ \hline
ELAN-light\cite{ref51}                                                             & ×3                                                                   & 75.7                                                                                          & 105                                                                                              & 590                                                                                            & \multicolumn{1}{l|}{\cellcolor[HTML]{FFFFFF}34.61}          & 0.9288                                            & \multicolumn{1}{l|}{\cellcolor[HTML]{FFFFFF}\underline{30.55}}          & \underline{0.8463}                                            & \multicolumn{1}{l|}{\cellcolor[HTML]{FFFFFF}29.21}          & 0.8081                                            & \multicolumn{1}{l|}{\cellcolor[HTML]{FFFFFF}\underline{28.69}}          & \underline{0.8624}                                            & \multicolumn{1}{l|}{\cellcolor[HTML]{FFFFFF}34.00}          & \underline{0.9478}                                            \\ \hline
SwinIR-light\cite{ref12}                                                           & ×3                                                                   & 87.2                                                                                          & 445                                                                                              & 886                                                                                            & \multicolumn{1}{l|}{\cellcolor[HTML]{FFFFFF}34.62}          & \underline{0.9289}                                            & \multicolumn{1}{l|}{\cellcolor[HTML]{FFFFFF}30.54}          & \underline{0.8463}                                            & \multicolumn{1}{l|}{\cellcolor[HTML]{FFFFFF}29.20}          & 0.8082                                            & \multicolumn{1}{l|}{\cellcolor[HTML]{FFFFFF}28.66}          & \underline{0.8624}                                            & \multicolumn{1}{l|}{\cellcolor[HTML]{FFFFFF}33.98}          & \underline{0.9478}                                            \\ \hline
SMFANET\cite{SMFANET }                                                               & ×3                                                                   & 48.0                                                                                          & -                                                                                                & 487                                                                                            & \multicolumn{1}{l|}{\cellcolor[HTML]{FFFFFF}\underline{34.63}}          & 0.9285                                            & \multicolumn{1}{l|}{\cellcolor[HTML]{FFFFFF}30.52}          & 0.8456                                            & \multicolumn{1}{l|}{\cellcolor[HTML]{FFFFFF}\underline{29.23}}          & \underline{0.8084}                                            & \multicolumn{1}{l|}{\cellcolor[HTML]{FFFFFF}28.59}          & 0.8594                                            & \multicolumn{1}{l|}{\cellcolor[HTML]{FFFFFF}\underline{34.17}}          & \underline{0.9478}                                            \\ \hline
ESTN (ours)                                                            & ×3                                                                   & 125.5                                                                                         & 335                                                                                              & 871                                                                                            & \multicolumn{1}{l|}{\cellcolor[HTML]{FFFFFF}\textbf{\begin{tabular}[c]{@{}l@{}}34.68\\ 0.05\%↑\end{tabular}}} & \textbf{\begin{tabular}[c]{@{}l@{}}0.9298\\ 0.10\%↑\end{tabular}}                                   & \multicolumn{1}{l|}{\cellcolor[HTML]{FFFFFF}\textbf{\begin{tabular}[c]{@{}l@{}}30.61\\ 0.20\%↑\end{tabular}}} & \textbf{\begin{tabular}[c]{@{}l@{}}0.8476\\ 0.15\%↑\end{tabular}}                                   & \multicolumn{1}{l|}{\cellcolor[HTML]{FFFFFF}\textbf{\begin{tabular}[c]{@{}l@{}}29.25\\ 0.07\%↑\end{tabular}}} & \textbf{\begin{tabular}[c]{@{}l@{}}0.8096\\ 0.15\%↑\end{tabular}}                                   & \multicolumn{1}{l|}{\cellcolor[HTML]{FFFFFF}\textbf{\begin{tabular}[c]{@{}l@{}}28.82\\ 0.45\%↑\end{tabular}}} & \textbf{\begin{tabular}[c]{@{}l@{}}0.8661\\ 0.43\%↑\end{tabular}}                                   & \multicolumn{1}{l|}{\cellcolor[HTML]{FFFFFF}\textbf{\begin{tabular}[c]{@{}l@{}}34.28\\ 0.32\%↑\end{tabular}}} & \textbf{\begin{tabular}[c]{@{}l@{}}0.9491\\ 0.14\%↑\end{tabular}}                                   \\ \hline
SRCNN\cite{ref8}                                                                  & ×4                                                                   & -                                                                                             & -                                                                                                & 57                                                                                             & \multicolumn{1}{l|}{\cellcolor[HTML]{FFFFFF}30.48}          & 0.8628                                            & \multicolumn{1}{l|}{\cellcolor[HTML]{FFFFFF}27.49}          & 0.7503                                            & \multicolumn{1}{l|}{\cellcolor[HTML]{FFFFFF}25.90}          & 0.7101                                            & \multicolumn{1}{l|}{\cellcolor[HTML]{FFFFFF}24.52}          & 0.7221                                            & \multicolumn{1}{l|}{\cellcolor[HTML]{FFFFFF}27.66}          & 0.8505                                            \\ \hline
CARN\cite{ref47}                                                                   & ×4                                                                   & 90.9                                                                                          & 30                                                                                               & 1592                                                                                           & \multicolumn{1}{l|}{\cellcolor[HTML]{FFFFFF}32.13}          & 0.8937                                            & \multicolumn{1}{l|}{\cellcolor[HTML]{FFFFFF}28.60}          & 0.7806                                            & \multicolumn{1}{l|}{\cellcolor[HTML]{FFFFFF}27.58}          & 0.7349                                            & \multicolumn{1}{l|}{\cellcolor[HTML]{FFFFFF}26.07}          & 0.7837                                            & \multicolumn{1}{l|}{\cellcolor[HTML]{FFFFFF}30.47}          & 0.9084                                            \\ \hline
IMDN\cite{ref48}                                                                   & ×4                                                                   & 40.9                                                                                          & 19                                                                                               & 715                                                                                            & \multicolumn{1}{l|}{\cellcolor[HTML]{FFFFFF}32.21}          & 0.8948                                            & \multicolumn{1}{l|}{\cellcolor[HTML]{FFFFFF}28.58}          & 0.7813                                            & \multicolumn{1}{l|}{\cellcolor[HTML]{FFFFFF}27.56}          & 0.7353                                            & \multicolumn{1}{l|}{\cellcolor[HTML]{FFFFFF}26.04}          & 0.7838                                            & \multicolumn{1}{l|}{\cellcolor[HTML]{FFFFFF}30.45}          & 0.9075                                            \\ \hline
LAPAR-A\cite{ref49}                                                                & ×4                                                                   & 94.0                                                                                          & 47                                                                                               & 659                                                                                            & \multicolumn{1}{l|}{\cellcolor[HTML]{FFFFFF}32.15}          & 0.8944                                            & \multicolumn{1}{l|}{\cellcolor[HTML]{FFFFFF}28.61}          & 0.7818                                            & \multicolumn{1}{l|}{\cellcolor[HTML]{FFFFFF}27.61}          & 0.7366                                            & \multicolumn{1}{l|}{\cellcolor[HTML]{FFFFFF}26.14}          & 0.7871                                            & \multicolumn{1}{l|}{\cellcolor[HTML]{FFFFFF}30.42}          & 0.9074                                            \\ \hline
ESRT\cite{ref50}                                                                   & ×4                                                                   & -                                                                                             & -                                                                                                & 751                                                                                            & \multicolumn{1}{l|}{\cellcolor[HTML]{FFFFFF}32.19}          & 0.8947                                            & \multicolumn{1}{l|}{\cellcolor[HTML]{FFFFFF}28.69}          & 0.7833                                            & \multicolumn{1}{l|}{\cellcolor[HTML]{FFFFFF}27.69}          & 0.7379                                            & \multicolumn{1}{l|}{\cellcolor[HTML]{FFFFFF}26.39}          & 0.7962                                            & \multicolumn{1}{l|}{\cellcolor[HTML]{FFFFFF}30.75}          & 0.9100                                            \\ \hline
ELAN-light\cite{ref51}                                                             & ×4                                                                   & 43.2                                                                                          & 62                                                                                               & 601                                                                                            & \multicolumn{1}{l|}{\cellcolor[HTML]{FFFFFF}32.43}          & 0.8975                                            & \multicolumn{1}{l|}{\cellcolor[HTML]{FFFFFF}\underline{28.78}}          & \underline{0.7858}                                            & \multicolumn{1}{l|}{\cellcolor[HTML]{FFFFFF}27.69}          & \underline{0.7406}                                            & \multicolumn{1}{l|}{\cellcolor[HTML]{FFFFFF}\underline{26.47}}          & \underline{0.7982}                                            & \multicolumn{1}{l|}{\cellcolor[HTML]{FFFFFF}30.92}          & \underline{0.9150}                                            \\ \hline
SwinIR-light\cite{ref12}                                                           & ×4                                                                   & 49.6                                                                                          & 271                                                                                              & 897                                                                                            & \multicolumn{1}{l|}{\cellcolor[HTML]{FFFFFF}\underline{32.44}}          & 0.8975                                            & \multicolumn{1}{l|}{\cellcolor[HTML]{FFFFFF}28.77}          & \underline{0.7858}                                            & \multicolumn{1}{l|}{\cellcolor[HTML]{FFFFFF}27.69}          & \underline{0.7406}                                            & \multicolumn{1}{l|}{\cellcolor[HTML]{FFFFFF}\underline{26.47}}          & 0.7980                                            & \multicolumn{1}{l|}{\cellcolor[HTML]{FFFFFF}30.92}          & \underline{0.9150}                                            \\ \hline
DIPNet\cite{DIPNet}                                                                 & ×4                                                                   & -                                                                                             & -                                                                                                & 543                                                                                            & \multicolumn{1}{l|}{\cellcolor[HTML]{FFFFFF}32.20}          & 0.8950                                            & \multicolumn{1}{l|}{\cellcolor[HTML]{FFFFFF}28.58}          & 0.7811                                            & \multicolumn{1}{l|}{\cellcolor[HTML]{FFFFFF}27.59}          & 0.7364                                            & \multicolumn{1}{l|}{\cellcolor[HTML]{FFFFFF}26.16}          & 0.7879                                            & \multicolumn{1}{l|}{\cellcolor[HTML]{FFFFFF}30.53}          & 0.9087                                            \\ \hline
SMFANET\cite{SMFANET }                                                               & ×4                                                                   & 28.0                                                                                          & -                                                                                                & 496                                                                                            & \multicolumn{1}{l|}{\cellcolor[HTML]{FFFFFF}32.43}          & \underline{0.8979}                                            & \multicolumn{1}{l|}{\cellcolor[HTML]{FFFFFF}28.77}          & 0.7849                                            & \multicolumn{1}{l|}{\cellcolor[HTML]{FFFFFF}\underline{27.70}}          & 0.7400                                            & \multicolumn{1}{l|}{\cellcolor[HTML]{FFFFFF}26.45}          & 0.7943                                            & \multicolumn{1}{l|}{\cellcolor[HTML]{FFFFFF}\underline{31.06}}          & 0.9138                                            \\ \hline
ESTN (ours)                                                            & ×4                                                                   & 75.1                                                                                          & 202                                                                                              & 881                                                                                            & \multicolumn{1}{l|}{\cellcolor[HTML]{FFFFFF}\textbf{\begin{tabular}[c]{@{}l@{}}32.55\\ 0.40\%↑\end{tabular}}} & \textbf{\begin{tabular}[c]{@{}l@{}}0.8993\\ 0.17\%↑\end{tabular}}                                   & \multicolumn{1}{l|}{\cellcolor[HTML]{FFFFFF}\textbf{\begin{tabular}[c]{@{}l@{}}28.83\\ 0.17\%↑\end{tabular}}} & \textbf{\begin{tabular}[c]{@{}l@{}}0.7876\\ 0.23\%↑\end{tabular}}                                   & \multicolumn{1}{l|}{\cellcolor[HTML]{FFFFFF}\textbf{\begin{tabular}[c]{@{}l@{}}27.71\\ 0.04\%↑\end{tabular}}} & \textbf{\begin{tabular}[c]{@{}l@{}}0.7421\\ 0.20\%↑\end{tabular}}                                   & \multicolumn{1}{l|}{\cellcolor[HTML]{FFFFFF}\textbf{\begin{tabular}[c]{@{}l@{}}26.67\\ 0.77\%↑\end{tabular}}} & \textbf{\begin{tabular}[c]{@{}l@{}}0.8040\\ 0.73\%↑\end{tabular}}                                   & \multicolumn{1}{l|}{\cellcolor[HTML]{FFFFFF}\textbf{\begin{tabular}[c]{@{}l@{}}31.13\\ 0.23\%↑\end{tabular}}} & \textbf{\begin{tabular}[c]{@{}l@{}}0.9166\\ 0.18\%↑\end{tabular}}                                   \\ \hline
\end{tabular}
}
\end{center}
\end{table*}

\begin{table}
  \centering
  \caption{Module Ablation Experiment Manga109 Dataset at 4× Upscaling\label{tab:table2}}
    \begin{tabular}{|c|c|c|c|}
    \hline
     \multirow{2}{*}{Method} & \multirow{2}{*}{ELAN-light} & \multirow{2}{*}{+BSGM} & +BSGM \\
            &            &       &+LRCAB \\
    \hline
    FLOPs & 54G & 71G & 75G \\
    \hline
    Params & 601k & 715k & 883k \\
    \hline
    PSNR (dB) & 30.67 & 30.78 & 30.87 \\
    \hline
    SSIM  & 0.9112 & 0.9120 & 0.9128 \\
    \hline
    \end{tabular}
\end{table}

\subsection{Experimental Setup}
The quality of SR reconstruction is assessed using Peak Signal-to-Noise Ratio\cite{ref45} (PSNR) and Structural Similarity\cite{ref46} (SSIM). Higher PSNR and SSIM (closer to 1) indicate superior image quality and greater structural similarity between SR and HR, respectively, with PSNR and SSIM expressed in Eqs. \eqref{eq-PSNR} and \eqref{eq-SSIM}.
\begin{equation}
\label{eq-PSNR}
{PSNR = 20 \cdot {\log _{10}}\left( {\frac{{{2^n} - 1}}{{RMSE\left( {{\boldsymbol{I}^{SR}},{\boldsymbol{I}^{HR}}} \right)}}} \right),}
\end{equation}\\
where $RMSE$ represents the root mean square error operation operator of the image; $n$ denotes the number of bits in the image; $\boldsymbol{I}^{SR}$ and $\boldsymbol{I}^{HR}$ signify SR reconstructed images and original HR images, respectively.
\begin{small}
\begin{equation}
\label{eq-SSIM}
SSIM = \frac{{\left( {2{{\boldsymbol{\overline I} }^{HR}} \cdot {{\boldsymbol{\overline I} }^{SR}} + {a_1}} \right)\left( {2{\sigma _{{\boldsymbol{I}^{HR}}{\boldsymbol{I}^{SR}}}} + {a_2}} \right)}}{{\left( {{{\left( {{{\boldsymbol{\overline I} }^{HR}}} \right)}^2} + {{\left( {{{\boldsymbol{\overline I} }^{SR}}} \right)}^2} + {a_1}} \right)  \left( {{{\left( {{\sigma _{{\boldsymbol{I}^{HR}}}}} \right)}^2} + {{\left( {{\sigma _{{\boldsymbol{I}^{SR}}}}} \right)}^2} + {a_2}} \right)}},
\end{equation}
\end{small}\\
where $\boldsymbol{\overline I}$ represents the average grayscale value of the image; $\sigma$ refers to the standard deviation of the image; ${\sigma _{{\boldsymbol{I}^{HR}}{\boldsymbol{I}^{SR}}}}$ denotes the covariance of $\boldsymbol{I}^{SR}$ and $\boldsymbol{I}^{HR}$; $a_1$ and $a_2$ signify constant coefficients determined by the image's pixel value range.
\subsubsection{Training Details}
\ 

The proposed network is trained on the DIV2K SR dataset with 800 LR-HR image pairs. The HR images are cropped to 256×256, with a mini-batch data size of $N = 64$. Five test datasets are employed for comparison: Set5\cite{ref40}, Set14\cite{ref41}, BSD100\cite{ref42}, Urban100\cite{ref43}, and Manga109\cite{ref44}.
\subsubsection{Training Setup}
\ 

\begin{figure*}[!t]
    \centering
    \subfigure[HR]{
        \centering
        \includegraphics[width=2.2in]{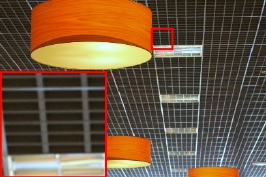}
    }
    \subfigure[LR]{
        \centering
        \includegraphics[width=2.2in]{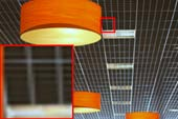}
    }
    \subfigure[CARN]{
        \centering
        \includegraphics[width=2.2in]{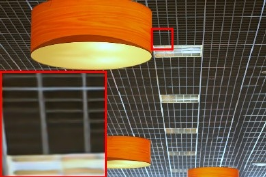}
    }
    \subfigure[IMDN]{
        \centering
        \includegraphics[width=2.2in]{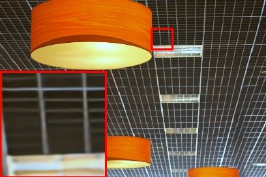}
    }
    \subfigure[LAPAR-A]{
        \centering
        \includegraphics[width=2.2in]{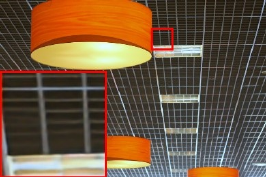}
    }
    \subfigure[ESRT]{
        \centering
        \includegraphics[width=2.2in]{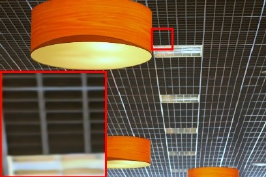}
    }
        \subfigure[ELAN-light]{
        \centering
        \includegraphics[width=2.2in]{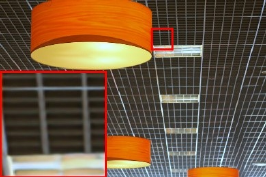}
    }
    \subfigure[SwinIR-light]{
        \centering
        \includegraphics[width=2.2in]{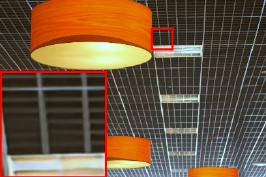}
    }
    \subfigure[Ours]{
        \centering
        \includegraphics[width=2.2in]{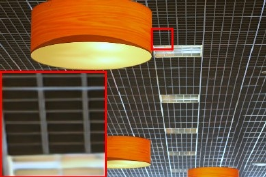}
    }
    \caption{Qualitative comparison of img044 with state-of-the-art lightweight SR models at x4 scale. ESTN demonstrates excellent performance in reconstructing clearer and sharper edge textures than other models.}
\label{fig_10}
\end{figure*}
\begin{figure*}[!t]
    \centering
    \subfigure[HR]{
        \centering
        \includegraphics[width=2.2in]{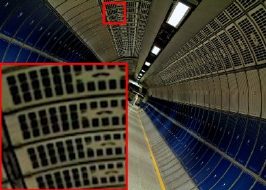}
    }
    \subfigure[LR]{
        \centering
        \includegraphics[width=2.2in]{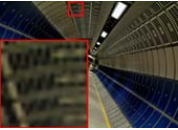}
    }
    \subfigure[CARN]{
        \centering
        \includegraphics[width=2.2in]{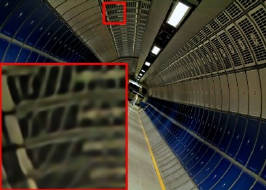}
    }
    \subfigure[IMDN]{
        \centering
        \includegraphics[width=2.2in]{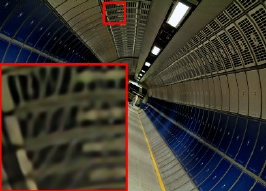}
    }
    \subfigure[LAPAR-A]{
        \centering
        \includegraphics[width=2.2in]{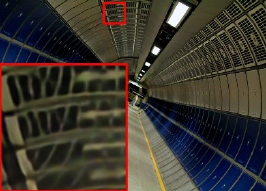}
    }
    \subfigure[ESRT]{
        \centering
        \includegraphics[width=2.2in]{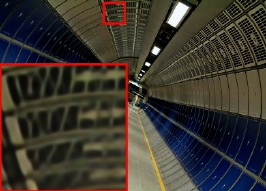}
    }
        \subfigure[ELAN-light]{
        \centering
        \includegraphics[width=2.2in]{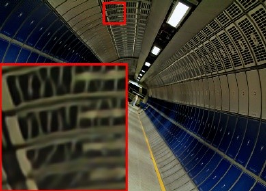}
    }
    \subfigure[SwinIR-light]{
        \centering
        \includegraphics[width=2.2in]{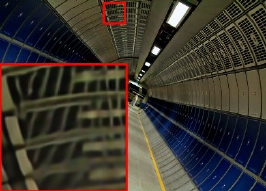}
    }
    \subfigure[Ours]{
        \centering
        \includegraphics[width=2.2in]{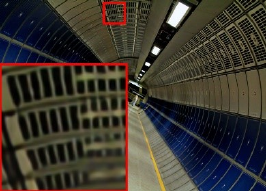}
    }
    \caption{Qualitative comparison of img078 with state-of-the-art lightweight SR models at a 4× scale. ESTN achieves more accurate texture reconstruction than other models.}
\label{fig_11}
\end{figure*}
\begin{figure*}[!t]
    \centering
    \subfigure[HR]{
        \centering
        \includegraphics[width=2.2in]{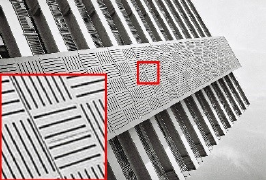}
    }
    \subfigure[LR]{
        \centering
        \includegraphics[width=2.2in]{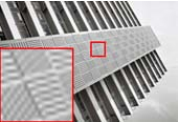}
    }
    \subfigure[CARN]{
        \centering
        \includegraphics[width=2.2in]{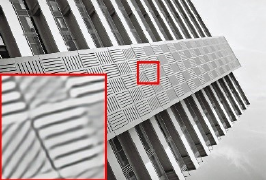}
    }
    \subfigure[IMDN]{
        \centering
        \includegraphics[width=2.2in]{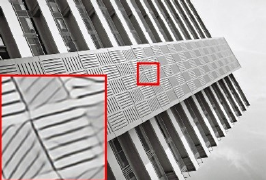}
    }
    \subfigure[LAPAR-A]{
        \centering
        \includegraphics[width=2.2in]{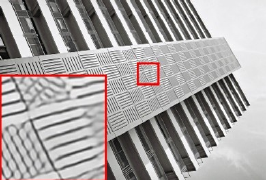}
    }
    \subfigure[ESRT]{
        \centering
        \includegraphics[width=2.2in]{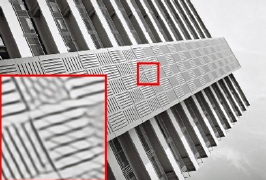}
    }
        \subfigure[ELAN-light]{
        \centering
        \includegraphics[width=2.2in]{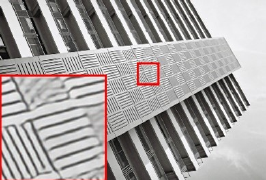}
    }
    \subfigure[SwinIR-light]{
        \centering
        \includegraphics[width=2.2in]{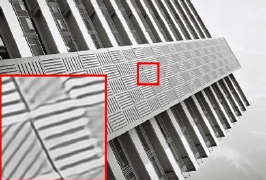}
    }
    \subfigure[Ours]{
        \centering
        \includegraphics[width=2.2in]{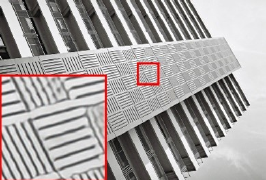}
    }
    \caption{Qualitative comparison of img092 with state-of-the-art lightweight SR models at a 4× scale. The proposed ESTN network demonstrates remarkable reconstruction of comprehensive and accurate edge information compared to other models.}
\label{fig_12}
\end{figure*}

This study conducts ×2, ×3, and ×4 upscale SR reconstruction tasks during training. The proposed ESTN comprises 12 ESTM blocks, each with channel numbers $C = 60$. The BSGM window is set at 4×4, while the W/SW-MSSA module's multiscale windows are set to 4×4, 8×8, and 16×16. To reduce computational overhead, the attention scores computed in W-MSSA are shared with SW-MSSA. Training image pairs for ESTN are generated via bicubic downsampling, and each batch comprises 64 randomly cropped image patches of size 64×64 from the LR images. The network is trained for 500 iterations with an initial learning rate of 0.0002, halved at the 250-th, 400-th, 425-th, 450-th, and 475-th iterations. The Adam optimizer is used based on ${\beta _1} = 0.9,{\beta _2} = 0.999$, and weight decay = 1e--8. All experiments were conducted on a server with two NVIDIA RTX3090 GPU cards.
\subsubsection{Test Setup}
\ 

We aim to enhance the model’s lightweight performance and reconstruction quality. The lightweight performance is determined by the number of parameters (Params) and float point operations (FLOPs), with FLOPs computed by upscaling the SR image resolution to 1280×720. The reconstruction quality is evaluated using the PSNR\cite{ref45} and SSIM\cite{ref46} indicators. The SR image is transformed from the RGB to YCbCr space, and the PSNR and SSIM are computed on the Y channel.

\subsection{Comparison with State-of-the-Art Models}
This study compares the ESTN against seven state-of-the-art single-image SR lightweight SR models: SRCNN\cite{ref8}, CARN\cite{ref47}, IMDN\cite{ref48}, LAPAR-A\cite{ref49}, ESRT\cite{ref50}, ELAN-light\cite{ref51} , and SwinIR-light\cite{ref12}. The experimental data is derived from weight parameters or SR results. Since ELAN-light\cite{ref51} only offers the source code, we train and obtain the results for comparison.
\subsubsection{Quantitative Comparison}
\ 

As presented in Table \ref{tab:table1}, the proposed ESTN demonstrates state-of-the-art performance in SR reconstruction across all five test sets. In the ×4 upscale SR results, ESTN performs robustly, even on the challenging Urban100 and Manga109 datasets. The Manga109 achieves a PSNR improvement of 0.21 dB over ELAN-light\cite{ref51} and SwinIR-light\cite{ref12}. Moreover, ESTN achieves outstanding performance improvements on Set5, Set14, BSD100, and Urban100 datasets. The ESTN contains fewer parameters and performs better than SwinIR-light\cite{ref12}.

\subsubsection{Qualitative Comparison}
\ 

The qualitative comparison of ×4 SR results on img044, img078, and img092 images in Urban100 is shown in Figs. \ref{fig_10}, \ref{fig_11}, and \ref{fig_12}. Fig. \ref{fig_10} reveals that the enlarged part of the SR images produced by CNN-based models such as CARN\cite{ref47}, IMDN\cite{ref48}, and LAPAR-A\cite{ref49} exhibit significant blurring and poor visual effects. Although the ESRT\cite{ref50}, ELAN-light\cite{ref51}, and SwinIR-light\cite{ref12} models effectively preserve the texture of the images in SR images, edge blurring persisted. In contrast, ESTN reconstructs SR images with clear and sharp edges. Fig. \ref{fig_11} demonstrates that only the ESTN accurately restores the texture in the magnified part of the SR image, while other SR models, such as CARN\cite{ref47}, IMDN\cite{ref48}, LAPAR-A\cite{ref49}, ESRT\cite{ref50}, ELAN-light\cite{ref51}, SwinIR-light\cite{ref12}, DIPNet\cite{DIPNet}, and SMFANET \cite{SMFANET}, fail to recover the correct texture. In addition, only the ESTN effectively restores image textures across multiple directions in the enlarged SR region (Fig. \ref{fig_12}). In contrast, CNN-based methods of CARN\cite{ref47}, IMDN\cite{ref48}, and LAPAR-A\cite{ref49} produce incorrect and blurred direction textures. The ESRT\cite{ref50}, ELAN-light\cite{ref51}, and SwinIR-light\cite{ref12} fail to capture multiple directional texture details simultaneously.

The CNN-based SR models showed limited reconstruction quality due to CNN's small receptive fields. Although Transformer-based and Swin Transformer-based models offer performance enhancements over CNNs, they remain suboptimal. The proposed model addresses this limitation by expanding the receptive field through sparse global perception, thereby enabling a global receptive field. As a result, it achieves more accurate texture restoration than conventional Swin Transformer-based SR models.

The qualitative and quantitative analyses demonstrate that the proposed model outperforms other advanced methods. The SR images reconstructed by ESTN more closely match the HR image than those generated by alternative networks.

\begin{figure*}[!t]
    \centering
    \subfigure[LR]{
        \centering
        \includegraphics[height=1.9in]{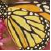}
    }
    \subfigure[ELAN-light S1]{
        \centering
        \includegraphics[height=1.9in]{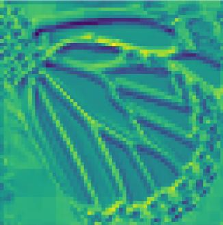}
    }
    \subfigure[ELAN-light S2]{
        \centering
        \includegraphics[height=1.9in]{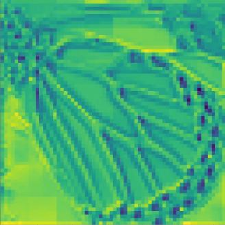}
    }
    \subfigure[HR]{
        \centering
        \includegraphics[height=1.9in]{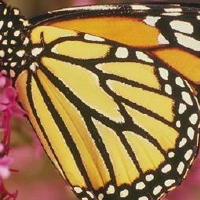}
    }
    \subfigure[ESTN S1]{
        \centering
        \includegraphics[height=1.9in]{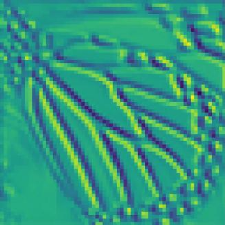}
    }
    \subfigure[ESTN S2]{
        \centering
        \includegraphics[height=1.9in]{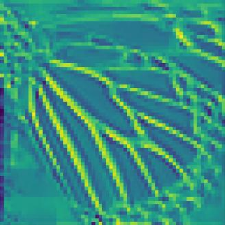}
    }

    \caption{Output features of the W/SW-SA modules for Butterfly images.}
\label{butt}
\end{figure*}

\begin{figure*}[!t]
    \centering
    \subfigure[LR]{
        \centering
        \includegraphics[height=1.9in]{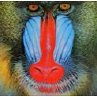}
    }
    \subfigure[ELAN-light S1]{
        \centering
        \includegraphics[height=1.9in]{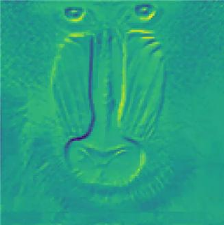}
    }
    \subfigure[ELAN-light S2]{
        \centering
        \includegraphics[height=1.9in]{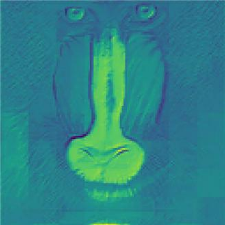}
    }
    \subfigure[HR]{
        \centering
        \includegraphics[height=1.9in]{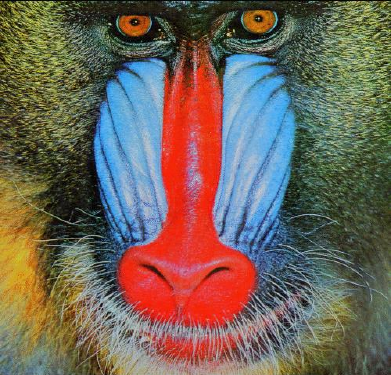}
    }
    \subfigure[ESTN S1]{
        \centering
        \includegraphics[height=1.9in]{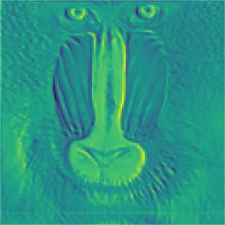}
    }
    \subfigure[ESTN S2]{
        \centering
        \includegraphics[height=1.9in]{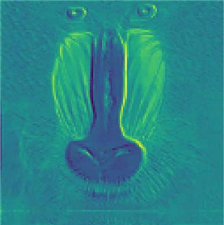}
    }

    \caption{Output features of the W/SW-SA modules for Baboon images.}
\label{babo}
\end{figure*}

\begin{figure*}[!t]
    \centering
    \subfigure[]{
        \centering
        \includegraphics[width=1.1in]{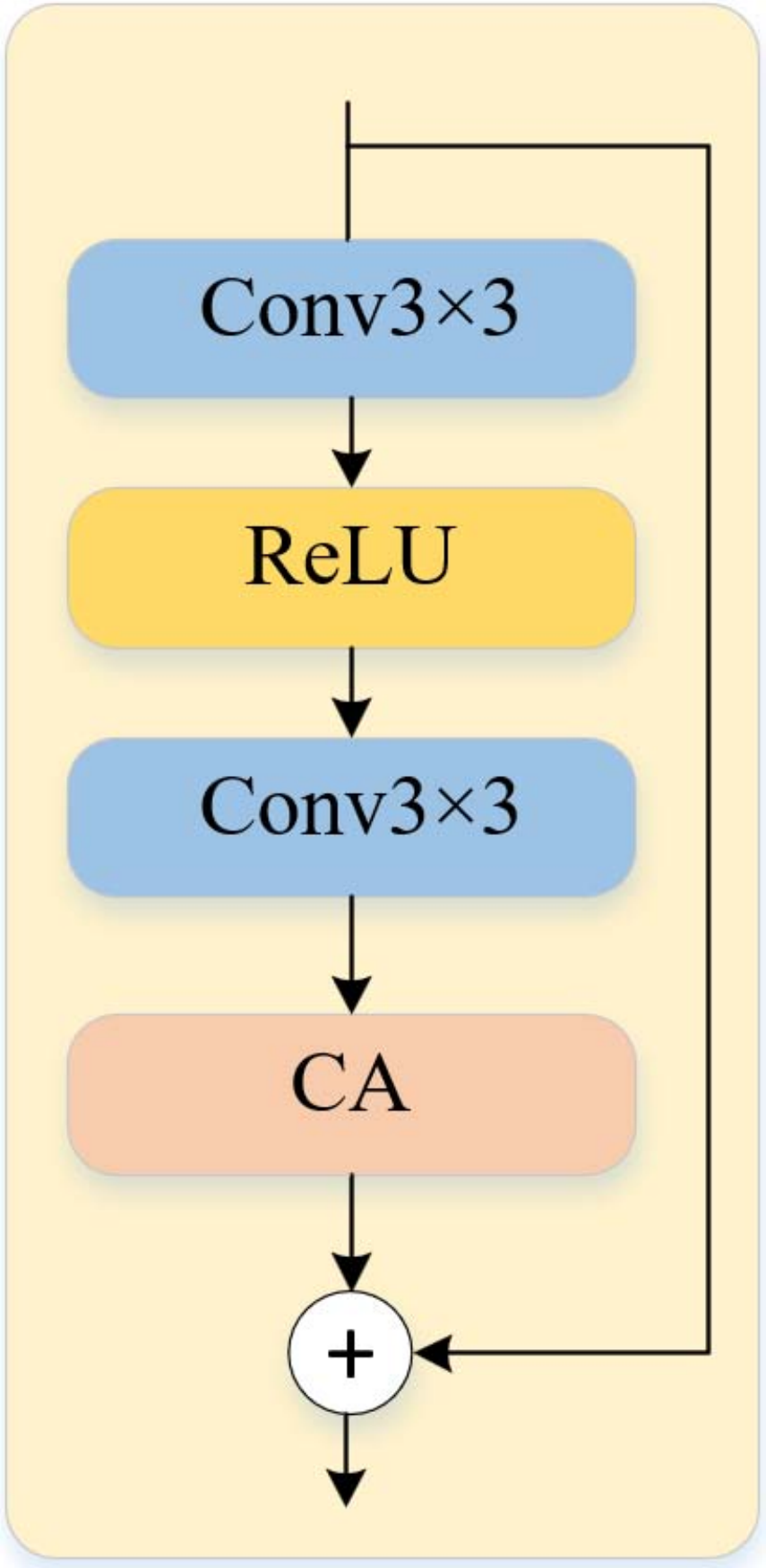}
    }
    \subfigure[]{
        \centering
        \includegraphics[width=1.1in]{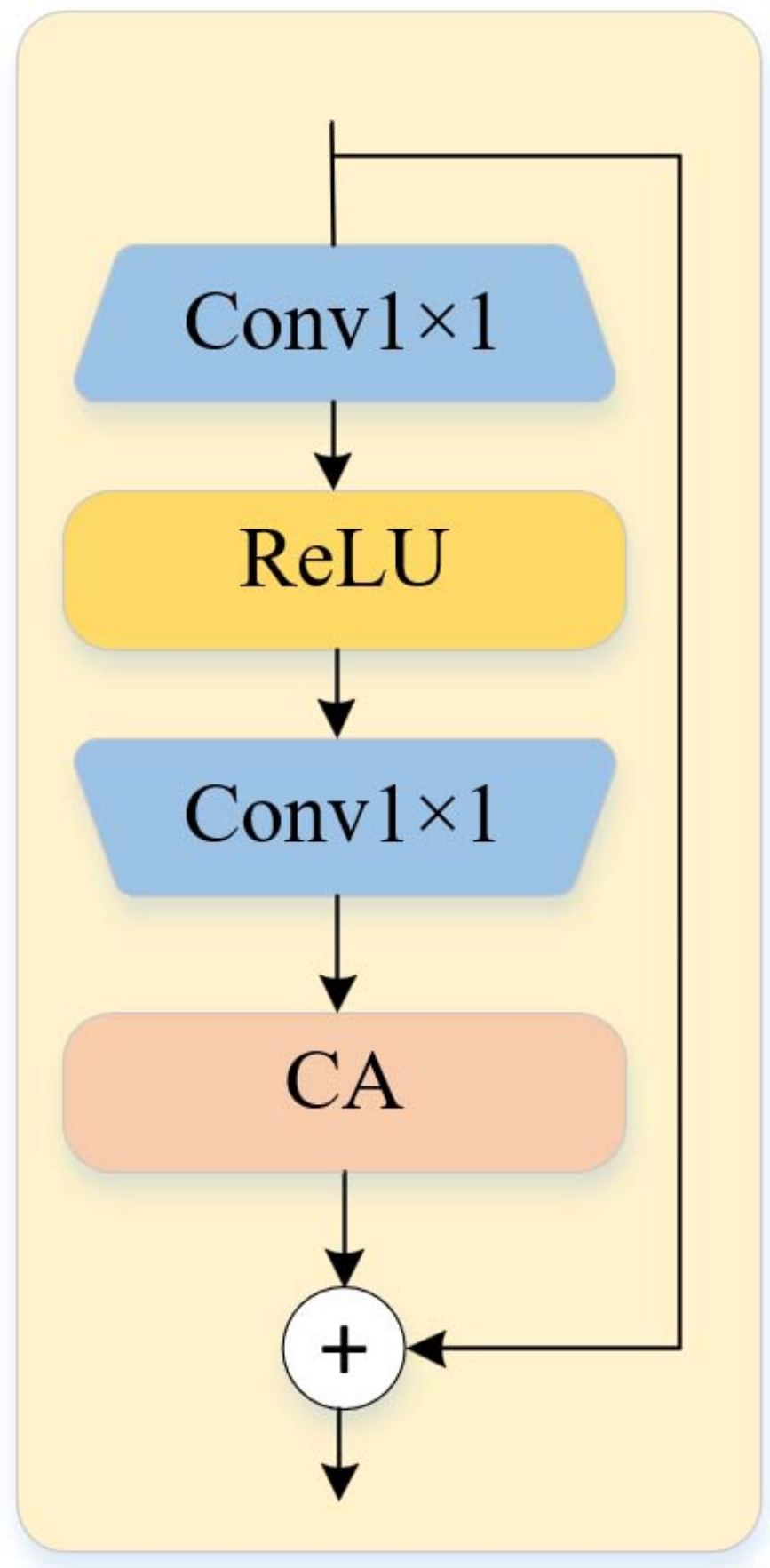}
    }
    \subfigure[]{
        \centering
        \includegraphics[width=1.1in]{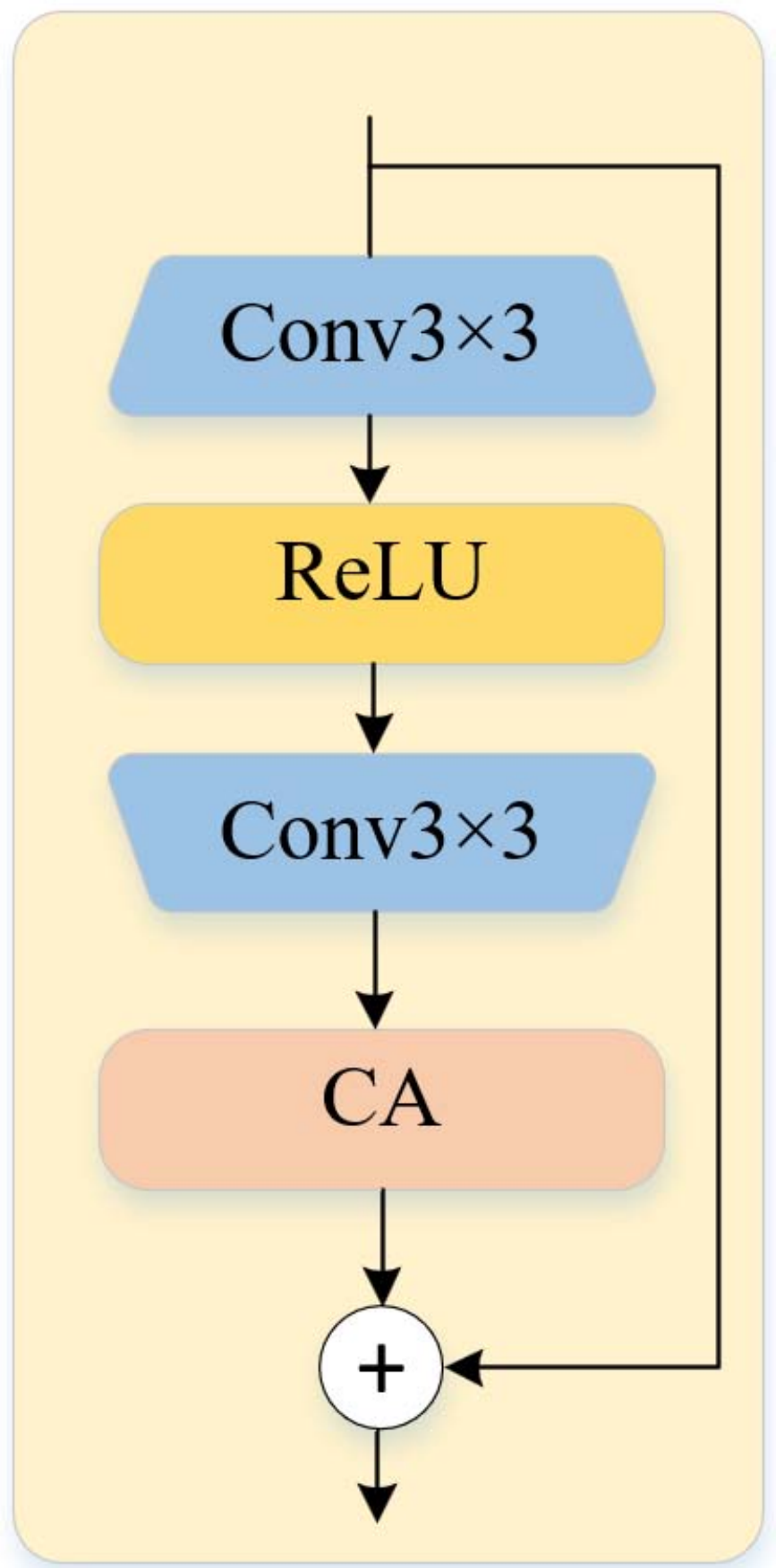}
    }
    \subfigure[]{
        \centering
        \includegraphics[width=1.1in]{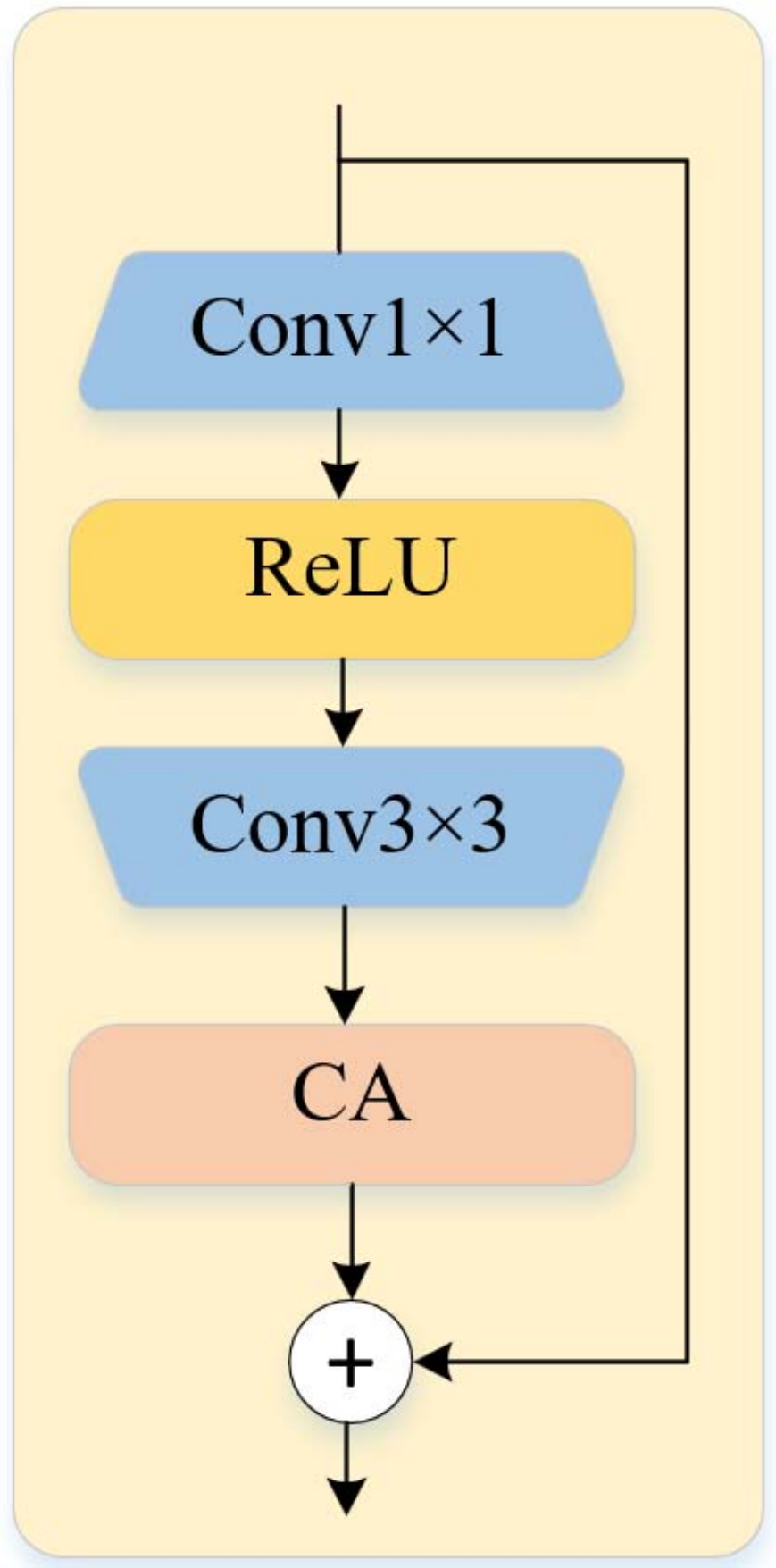}
    }
    \caption{Comparison of LRCAB with other types of channel attention modules. (a) Original RCAB. (b) RCAB with two 1×1 convolutions. (c) RCAB with two 3×3 convolutions. (d) RCAB with 1×1 \& 3×3 convolutions (LRCAB).}
\label{fig_15}
\end{figure*}

\begin{table*}
  \centering
  \caption{Ablation Study of the Channel Attention Module on the Manga109 Dataset at 4× Resolution\label{tab:table3}}
    \begin{tabular}{|c|c|c|c|c|}
    \hline
     \multirow{2}{*}{Design} & \multirow{2}{*}{Original RCAB} & \multirow{2}{*}{Two Conv1×1 RCAB} & \multirow{2}{*}{Two Conv3×3 RCAB} & Conv1×1 \& Conv3×3 \\
            &            &       &      & RCAB(LRCAB) \\
    \hline
    FLOPs & 71G & 79G & 72G & 75G\\
    \hline
    Params & 863k & 729k & 1036k & 883k \\
    \hline
    PSNR & 30.78 & 30.83 & 30.88 & 30.87\\
    \hline
    SSIM  & 0.9119 & 0.9119 & 0.9130 & 0.9128 \\
    \hline
    \end{tabular}
\end{table*}

Figs. \ref{butt} and \ref{babo} illustrate the feature maps produced by the window and shifted window self-attention modules in ELAN-light and ESTN, denoted as S1 and S2, respectively. The highlighted parts in the feature maps indicate the parts that are the model’s focus. In Fig. \ref{butt}, the S1 and S2 features output from ELAN-light shows limited attention to the image's textures. The S1 and S2 features generated by ESTN effectively enhance texture representation, resulting in clearer image reconstruction. In Fig. \ref{babo}, the S1 and S2 features from ELAN-light fail to capture fine texture and layering detail, while the S1 and S2 features from the ESTN preserve the texture part of the image, notably improving the clarity of the baboon’s beard texture.

\subsection{Ablation Studies}
We conducted ablation experiments to assess the contribution of each component of ESTN and various designs of low-parameter channel attention modules. All models in these experiments were trained with a batch size of 4, with other parameters consistent with the setup outlined in the experimental section.

\subsubsection{ESTN Ablation Studies}
\ 

The number of FLOPs and Params serve as reference metrics for measuring lightweight networks. To assess the effectiveness of the proposed strategy, we conducted ablation experiments with and without the BSGM and LRCAB. Table \ref{tab:table2} reveals that the network incorporating BSGM achieves a 0.12 dB improvement in PSNR over the ELAN-light\cite{ref51} network, with an increase of 114 K in the Params and 17 G in FLOPs. In contrast, the network with the LRCAB module achieves a 0.09 dB improvement in PSNR over ELAN-light\cite{ref51} network with BSGM, with an increase of 4 G in FLOPs and 168 K in Params. 

\subsubsection{Low-Parametric Residual Channel Attention Ablation Studies}
\ 

Figs. \ref{fig_15}(a)--(d) compare the efficiency of LRCAB across four channel attentions. Table \ref{tab:table3} demonstrates that the three redesigned channel attention blocks in Figs. \ref{fig_15}(b)--(d) significantly enhance the PSNR metrics relative to the Residual Channel Attention Block (RCAB). The module in Fig. \ref{fig_15}(b) applies two 1×1 convolutional transformations of the feature dimensions before computing channel attention. No performance improvement is observed, though it reduces the Params. The module in Fig. \ref{fig_15}(c) employs two 3×3 convolutions for feature transformation before channel attention, leading to improved performance metrics but an increase of 173 K in Params compared to the original RCAB. The LRCAB (Fig. \ref{fig_15}(d)) employs 1×1 and 3×3 convolutions to transform feature dimensions before computing the channel attention. This method enhances performance metrics and increases Params by 20 K and FLOPs by 4 G compared to the original RCAB.

\subsection{Analysis of Attribution Results}
\begin{figure*}[!t]
    \centering
    \includegraphics[width=0.8\textwidth]{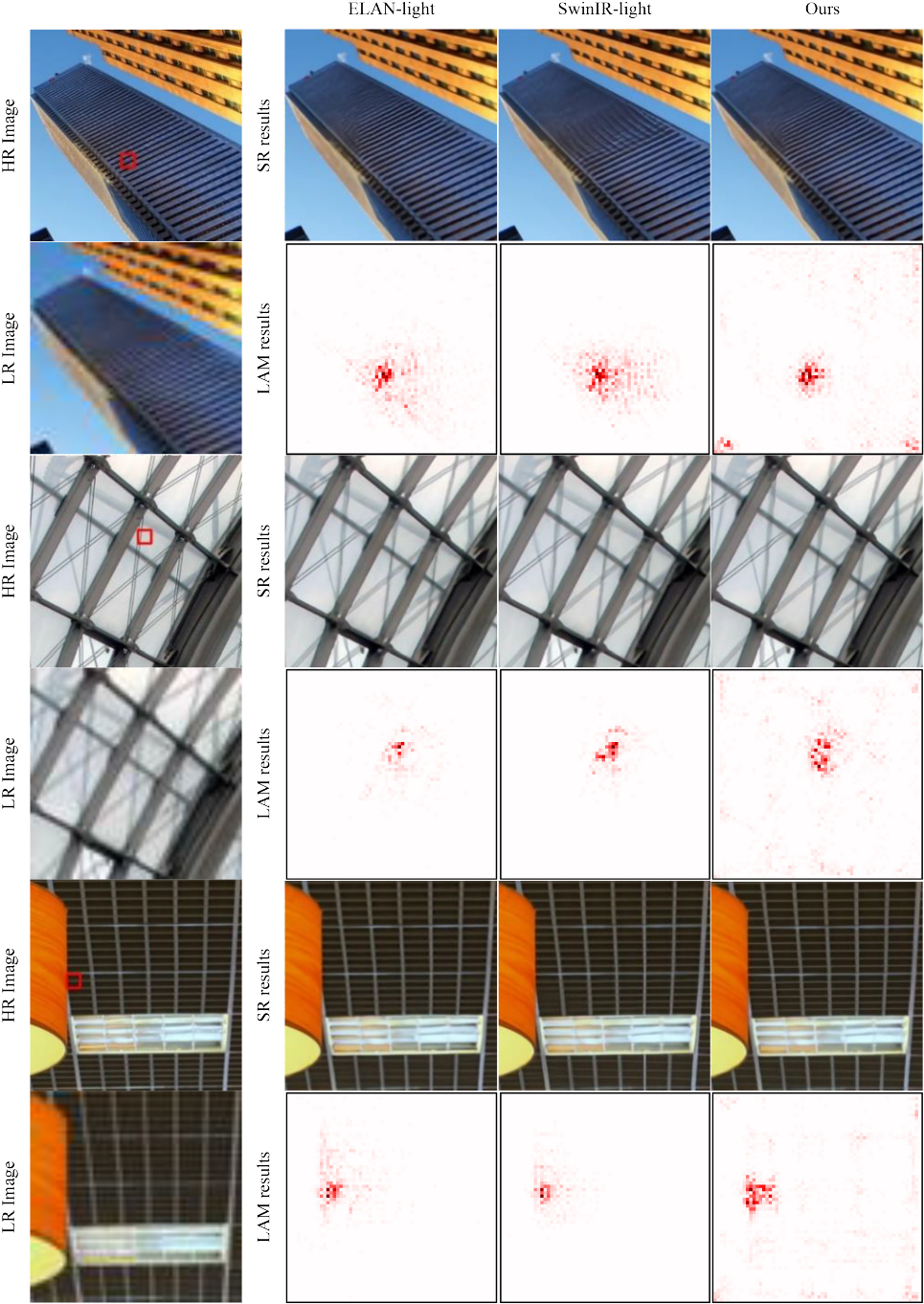}
    \caption{SR and LAM results for the Swin Transformer-based lightweight SR network, with LAM results visualizing the impact of different pixels on the SR output.}
\label{fig_16}    
\end{figure*}

Fig. \ref{fig_16} presents the SR and LAM results using the Transformer model. In the LAM results, red pixels highlight the significant influence on the recovery outcomes of the region of interest. For both ELAN-light\cite{ref51} and SwinIR-light\cite{ref12}, based on the Swin Transformer, the red pixels are predominantly concentrated around the selected region, indicating limited global perception capability. The LAM results for the ESTN model reveal that, apart from the dense red pixels near the selected area, red pixels are sparsely distributed across the entire region. Conclusively, the proposed model effectively utilizes information from the entire input LR image to restore the region of interest, enhancing image reconstruction and sharp texture restoration.

\section{Conclusion} 
This study proposes the image SR reconstruction network ESTN, which alternately aggregates local and global features to effectively enhance the network's receptive field and spatial-channel information exchange. The alternation of local-global feature aggregation fosters comprehensive spatial and channel interaction, improving the network’s nonlinear mapping capability. The shift convolution is introduced to aggregate local features and facilitate the local spatial-channel information interaction. In contrast, the BSGM, W-MSSA, SW-MSSA, and LRCAB enable global feature aggregation and the interaction between global spatial and channel information. The LAM results demonstrate that ESTN exhibits strong sparse global perception, confirming BSGM's ability to model sparse global information. By optimizing the RCAB structure for selective channel features, performance is enhanced without significantly increasing parameter count. Experimental results reveal that ESTN yields substantial performance improvements on the Set5, Set14, BSD100, Urban100, and Manga109 image SR datasets. 

Although the proposed model is significantly lightweight, it requires further weight reduction for SR reconstruction in practical applications. Deploying the model on edge devices may be impractical due to resource constraints. Furthermore, SR reconstruction in real-world settings is challenging, as images are influenced by various interferences that current models cannot adequately address. Introducing a generative adversarial mechanism could enhance SR reconstruction performance in such scenarios.

\section*{Acknowledgments}
This study is funded by the National Natural Science Foundation of China (Grant no. 62001199), the Natural Science Foundation Project of Fujian Province Grant nos. (2023J01155, 2024J01820, 2024J01821, and 2024J01822), and the Natural Science Foundation Project of Zhangzhou City (Grant no. ZZ2023J37). Additional support is provided by the Principal Foundation (Grant no. KJ19019), the High-Level Science Research Project (Grant no. GJ19019), and the Education Research Program (Grant no. 202211) of Minnan Normal University, and National Independent Innovation Demonstration Zone System (Fuzhou, Xiamen, Quanzhou) Innovation Platform Project (3502ZCQXT2024006), as well as the Undergraduate Education and Teaching Research Project of Fujian Province (Grant no. FBJY20230083). We thank LetPub (www.letpub.com.cn) for its linguistic assistance during the preparation of this manuscript.

\vfill
\vspace{-10 mm} 
\begin{IEEEbiography}[]{Yuming Huang}
	received the M.S. degree in Electronic Information from Minnan Normal University, Zhangzhou, Fujian, China, in 2024. He is currently a Laboratory technician with the School of Artificial Intelligence, Yulin Normal University, Yulin, Guangxi, China. His research interests include computer vision, deep learning and image processing.
\end{IEEEbiography}

\vspace{-10 mm} 
\begin{IEEEbiography}[]{Yingpin Chen}
	received his bachelor degree in Electronic Science and Technology from Fuzhou University, Fuzhou, Fujian, China, in 2009. He received his Ph.D. degree in Signal and Information Processing from University of Electronic and Science Technology of China, Chengdu, Sichuan, China, in 2019. He is currently an associate professor with Minnan Normal University, Zhangzhou, Fujian, China. His research interests include video object tracking, time-frequency analysis, and image processing. 
\end{IEEEbiography}

\vspace{-10 mm} 
\begin{IEEEbiography}[]{Changhui Wu}
	received his B.S. degree majoring in Electrical Engineering and Automation from Hanshan Normal University, Chaozhou, China, in 2022. He received the M.S. degree in Electronic Information from Minnan Normal University, Zhangzhou, Fujian, China, in 2025. His research interests include video object tracking and image processing.
\end{IEEEbiography}

\vspace{-10 mm} 
\begin{IEEEbiography}[]{Binhui Song}
	received his B.S. degree majoring in Electrical Engineering and Automation from Minnan Normal University, Zhangzhou, Fujian, China, in 2025. His research interests include video object tracking and image processing.
\end{IEEEbiography}

\vspace{-10 mm} 
\begin{IEEEbiography}[]{Hui Wang}
	received the Ph.D. degree majoring in communication and information system from Ningbo University, Ningbo, China, in 2019. He is currently an associate professor with the School of Physics and Information Engineering, Minnan Normal University, Zhangzhou, China. His current research interest includes wireless communication and network, machine learning and resource allocation.
\end{IEEEbiography} 

\end{document}